\documentclass[10pt,twocolumn,letterpaper]{article}

\usepackage{cvpr}
\usepackage{times}
\usepackage{epsfig}
\usepackage{graphicx}
\usepackage{xspace}
\usepackage{rotating}
\usepackage{multirow} 
\usepackage{cite}
\usepackage{url}
\usepackage{verbatim}
\usepackage{amsmath}
\usepackage{amssymb}
\usepackage{algorithmic}
\usepackage{tabularx}
\usepackage{subfig}
\cvprfinalcopy 

\def\cvprPaperID{2103} 

\begin{document}
\newcommand{\visionbank}{Visionbank~}

\newcommand{\Real}{\mathbb{R}} 

\newcommand{\abs}[1]{\ensuremath{\lvert#1\rvert}}
\newcommand{\norm}[1]{\ensuremath{\lVert#1\rVert}}
\newcommand{\Oh}{\mathcal{O}}
\newcommand{\oh}{\mathcal{o}}
\newcommand{\eat}{\ensuremath{\Big\vert_}}
\newcommand{\enh}[1]{\ensuremath{\tilde{#1}}}
\newcommand{\partials}[2]{\ensuremath{\frac{\partial#1}{\partial#2}}}
\newcommand{\lspan}[1]{\ensuremath{\mathrm{span}}}
\newcommand{\by}[2]{\ensuremath{#1 \! \times \! #2}}
\newcommand{\kl}[2]{\ensuremath{\mathcal{D}\left(#1 \! \parallel \! #2\right)}}

\newcommand{\trace}{\ensuremath{\mathop{\mathrm{tr}}}}
\newcommand{\diag}{\ensuremath{\mathop{\mathrm{diag}}}}
\newcommand{\blkdiag}{\ensuremath{\mathop{\mathrm{blkdiag}}}}
\newcommand{\argmin}{\ensuremath{\mathop{\mathrm{argmin}}}}
\newcommand{\argmax}{\ensuremath{\mathop{\mathrm{argmax}}}}
\newcommand{\Var}{\ensuremath{\mathop{\mathrm{Var}}}}
\newcommand{\NCC}{\ensuremath{\mathop{\mathrm{NCC}}}}

\newcommand{\normals}[2]{\ensuremath{\mathcal{N}(#1,#2)}}
\newcommand{\normal}[3]{\ensuremath{\mathcal{N}(#1;#2,#3)}}
\newcommand{\normalsi}[2]{\ensuremath{\mathcal{N}_I(#1,#2)}}
\newcommand{\normali}[3]{\ensuremath{\mathcal{N}_I(#1;#2,#3)}}

\newcommand{\lapls}[1]{\ensuremath{\mathcal{L}(#1)}}
\newcommand{\lapl}[2]{\ensuremath{\mathcal{L}(#1;#2)}}

\newcommand{\gumbels}[1]{\ensuremath{\mathcal{G}(#1)}}
\newcommand{\gumbel}[2]{\ensuremath{\mathcal{G}(#1;#2)}}

\newcommand{\expns}[1]{\ensuremath{\mathcal{E}(#1)}}
\newcommand{\expn}[2]{\ensuremath{\mathcal{E}(#1;#2)}}

\newcommand{\logistics}[2]{\ensuremath{\mathcal{LG}(#1,#2)}}
\newcommand{\logistic}[3]{\ensuremath{\mathcal{LG}(#1;#2,#3)}}

\newcommand{\logisticsz}[1]{\ensuremath{\mathcal{LG}(#1)}}
\newcommand{\logisticz}[2]{\ensuremath{\mathcal{LG}(#1;#2)}}

\newcommand{\mcomma}{\ensuremath{\,,}}
\newcommand{\mdot}{\ensuremath{\,.}}

\newcommand{\dwt}{\ensuremath{F_a}\xspace}      
\newcommand{\idwt}{\ensuremath{F_s}\xspace}     
\newcommand{\dwtd}{\ensuremath{\idwt^T}\xspace} 
\newcommand{\idwtd}{\ensuremath{\dwt^T}\xspace} 

\newcommand{\xs}{\ensuremath{x}}        %

\newcommand{\bv}{\ensuremath{\mathbf{b}}}        %
\newcommand{\qv}{\ensuremath{\mathbf{q}}}        %
\newcommand{\mv}{\ensuremath{\mathbf{m}}}        %
\newcommand{\nv}{\ensuremath{\mathbf{n}}}        %
\newcommand{\cv}{\ensuremath{\mathbf{c}}}        %
\newcommand{\fv}{\ensuremath{\mathbf{f}}}        %
\newcommand{\gv}{\ensuremath{\mathbf{g}}}        %
\newcommand{\hv}{\ensuremath{\mathbf{h}}}        %
\newcommand{\kv}{\ensuremath{\mathbf{k}}}        %
\newcommand{\pv}{\ensuremath{\mathbf{p}}}        %
\newcommand{\rv}{\ensuremath{\mathbf{r}}}        %
\newcommand{\sv}{\ensuremath{\mathbf{s}}}        %
\newcommand{\vv}{\ensuremath{\mathbf{v}}}        %
\newcommand{\wv}{\ensuremath{\mathbf{w}}}        %
\newcommand{\epsilonv}{\ensuremath{\bm{\epsilon}}}     %
\newcommand{\betav}{\ensuremath{\bm{\beta}}}     %
\newcommand{\gammav}{\ensuremath{\bm{\gamma}}}        %
\newcommand{\lambdav}{\ensuremath{\bm{\lambda}}}        %
\newcommand{\xiv}{\ensuremath{\bm{\xi}}}     %
\newcommand{\muv}{\ensuremath{\bm{\mu}}}     %
\newcommand{\nuv}{\ensuremath{\bm{\nu}}}     %
\newcommand{\zv}{\ensuremath{\mathbf{z}}}        %
\newcommand{\yv}{\ensuremath{\mathbf{y}}}        %
\newcommand{\xv}{\ensuremath{\mathbf{\xs}}}        %
\newcommand{\phiv}{\ensuremath{\bm{\phi}}}        %
\newcommand{\thetav}{\ensuremath{\bm{\theta}}}        %
\newcommand{\zetav}{\ensuremath{\bm{\zeta}}}        %
\newcommand{\omegav}{\ensuremath{\bm{\omega}}}        %
\newcommand{\piv}{\ensuremath{\bm{\pi}}}        %
\newcommand{\zerov}{\ensuremath{\mathbf{0}}}        %

\newcommand{\Am}{\ensuremath{\mathbf{A}}}        %
\newcommand{\Cm}{\ensuremath{\mathbf{C}}}        %
\newcommand{\Dm}{\ensuremath{\mathbf{D}}}        %
\newcommand{\Fm}{\ensuremath{\mathbf{F}}}        %
\newcommand{\Gm}{\ensuremath{\mathbf{G}}}        %
\newcommand{\Hm}{\ensuremath{\mathbf{H}}}        %
\newcommand{\Um}{\ensuremath{\mathbf{I}}}        %
\newcommand{\Km}{\ensuremath{\mathbf{K}}}        %
\newcommand{\Lm}{\ensuremath{\mathbf{L}}}        %
\newcommand{\Mm}{\ensuremath{\mathbf{M}}}        %
\newcommand{\Pm}{\ensuremath{\mathbf{P}}}        %
\newcommand{\Rm}{\ensuremath{\mathbf{R}}}        %
\newcommand{\Sm}{\ensuremath{\mathbf{S}}}        %
\newcommand{\Tm}{\ensuremath{\mathbf{T}}}        %
\newcommand{\Xm}{\ensuremath{\mathbf{X}}}        %
\newcommand{\Wm}{\ensuremath{\mathbf{W}}}        %
\newcommand{\Unitm}{\ensuremath{\mathbf{I}}}        %
\newcommand{\Info}{\ensuremath{J}}        %
\newcommand{\Infom}{\ensuremath{\mathbf{\Info}}}        %
\newcommand{\Gammam}{\ensuremath{\bm{\Gamma}}}        %
\newcommand{\Deltam}{\ensuremath{\bm{\Delta}}}        %
\newcommand{\Lambdam}{\ensuremath{\bm{\Lambda}}}        %
\newcommand{\Sigmam}{\ensuremath{\bm{\Sigma}}}        %

\newcommand{\mup}{\muv}
\newcommand{\Sigmap}{\Sigmam}
\newcommand{\kp}{\kv}
\newcommand{\Infop}{\Infom} 

\newcommand{\fig}{Fig.}
\newcommand{\tabl}{Table}
\newcommand{\sect}{Sec.}
\newcommand{\sects}{Secs.}
\newcommand{\eq}{Eq.}
\newcommand{\eqs}{Eqs.}

\newcommand{\expect}[1]{\ensuremath{\mathcal{E}\left\{#1\right\}}}
\newcommand{\expectt}[2]{\ensuremath{\mathcal{E}_{#1}\left\{#2\right\}}}
\newcommand{\expectf}[2]{\ensuremath{\mathcal{E}_{#1}\{#2\}}}      
\newcommand{\expectfp}[3]{\ensuremath{\mathcal{E}^{#1}_{#2}\{#3\}}} 

\newcommand{\reffig}[1]{Fig.~\ref{#1}}
\newcommand{\refsec}[1]{Sec.~\ref{#1}}
\newcommand{\refeq}[1]{Eq.~\ref{#1}}

\newcommand{\fix}{\marginpar{FIX}}
\newcommand{\new}{\marginpar{NEW}}
\newcommand{\vecmu}{\mbox{\boldmath $\mu$}}
\newcommand{\mycomment}[1]{}
\newcommand{\myemph}[1]{#1}

\newcommand{\reftable}[1]{Table.~\ref{#1}}
\newcommand{\figwh}[3]{\includegraphics[height=#1,width=#2]{\imroot #3}}
\newcommand{\figw}[2]{\includegraphics[width=#1]{\imroot #2}}
\newcommand{\figh}[2]{\includegraphics[height=#1]{\imroot #2}}

\newcommand{\ww}{\mathbf{W}}
\newcommand{\hh}{\mathbf{h}}
\newcommand{\wm}[1]{\mathbf{w}^{(#1)}}
\newcommand{\wmm}{\mathbf{w}}
\newcommand{\Loss}{\mathcal{L}}
\newcommand{\loss}{{l}}
\newcommand{\ff}{\mathbf{f}}

\newcommand{\subfigwd}[3]{\subfloat[][#3]{\putfigurew{#1}{#2}}}
\newcommand{\subfigwdhg}[4]{\subfloat[][#4]{\putfigurew{#1}{#2}{#3}}}
\newcommand{\nsubfigwd}[2]{\subfloat[][]{\includegraphics[width  = #1]{./#2}}}
\newcommand{\nsubfigwdhg}[3]{\subfloat[][]{\includegraphics[width  = #1,height = #2]{./#3}}}

\newcommand{\sfigwh}[4]{\subfloat[][]{\includegraphics[height=#1,width=#2]{\imroot #3}}}
\newcommand{\sfigw}[3]{\subfloat[][]{\includegraphics[width=#1]{\imroot #2}}}
\newcommand{\sfigh}[3]{\subfloat[][]{\includegraphics[height=#1]{\imroot #2}}}
\newcommand{\sol}{\mathcal{L}}
\newcommand{\parmp}{\mathbf{w^p}}
\newcommand{\parm}{\mathbf{w}}
\newcommand{\edgeset}{\mathcal{E}}
\newcommand{\parmu}{\mathbf{w^u}}
\newcommand{\score}{\mathcal{S}}
\newcommand{\im}{\mathcal{I}}

\newcommand{\ba}{\begin{eqnarray}}
\newcommand{\ea}{\end{eqnarray}}
\newcommand{\scln}{.17}
\newcommand{\sclhl}{.09}
\newcommand{\sclq}{.05}

\newcommand{\ubernet}{UberNet~}
\title{\ubernet: Training a `Universal' Convolutional Neural Network for Low-, Mid-, and High-Level Vision using Diverse Datasets and   Limited Memory}
	\author{Iasonas Kokkinos\\
	iasonas.kokkinos@ecp.fr\\
	CentraleSup\'elec - INRIA 
	}
	
	\sloppy
	\maketitle
	\thispagestyle{empty}
	
\begin{abstract}
In this work we  introduce a convolutional neural network (CNN) that jointly handles low-, mid-, and high-level vision tasks in a  unified architecture that is trained end-to-end.  Such a universal  network can act like a `swiss knife' for vision tasks;  we call this architecture an \ubernet to indicate its overarching nature. 
	
We address two main technical challenges that emerge when broadening up the range of tasks handled by a single CNN: (i) training a deep architecture while relying on diverse training sets and (ii) training many (potentially unlimited) tasks with a limited memory budget.
Properly addressing these two problems allows us to train  accurate predictors for a host of tasks, without compromising accuracy. 

Through these advances we train in an end-to-end manner a CNN that  simultaneously addresses (a) boundary detection (b) normal estimation (c) saliency estimation (d) semantic segmentation (e) human part segmentation (f) semantic boundary detection, (g) region proposal generation and object detection. We obtain competitive performance while jointly addressing all of these tasks in 0.7 seconds per frame on a single GPU. A demonstration  of this system can be found at \url{cvn.ecp.fr/ubernet/}.
\end{abstract}
\maketitle

\noindent

\newcommand{\rt}{.}
\newcommand{\wdtg}{.485\linewidth}

\newcommand{\ft}[1]{#1}
\newcommand{\resfigwdf}[1]
{
	{ \setlength{\tabcolsep}{.05em}	
		\begin{tabular}{cccc}
			\ft{Input} & \ft{Normals} & \ft{Normals} & \ft{Normals}\\
			\includegraphics[width=\wdtg]{\rt/l#1_detection}&
			\includegraphics[width=\wdtg]{\rt/l#1_boundaries}&	
			\includegraphics[width=\wdtg]{\rt/l#1_surface_normals}&
			\includegraphics[width=\wdtg]{\rt/l#1_saliency}\\
			\includegraphics[width=\wdtg]{\rt/l#1_detection}&
			\includegraphics[width=\wdtg]{\rt/l#1_semantic_bounaries}& 
			\includegraphics[width=\wdtg]{\rt/#1_semantic_segmentation}& 
			\includegraphics[width=\wdtg]{\rt/#1_human_parts}
		\end{tabular}
	}
}
\newcommand{\resfigwdt}[1]
{
	{\setlength{\tabcolsep}{.2em}	
		\begin{tabular}{cc}
			\ft{Input} & \ft{Boundaries} \\
			\includegraphics[width=\wdtg]{\rt/l#1_input}&
			\includegraphics[width=\wdtg]{\rt/l#1_boundaries}\\
			 \ft{Surface Normals} & \ft{Saliency}\\	
			\includegraphics[width=\wdtg]{\rt/l#1_surface_normals.png}&
			\includegraphics[width=\wdtg]{\rt/l#1_saliency}\\
			 \ft{Semantic Segmentation} & \ft{Semantic Boundaries}\\
			\includegraphics[width=\wdtg]{\rt/l#1_semantic_segmentation.png}&  
			\includegraphics[width=\wdtg]{\rt/l#1_semantic_boundaries.png}\\
			 \ft{Human Parts} &  \ft{Detection} \\
			\includegraphics[width=\wdtg]{\rt/l#1_human_parts.png} & 
			\includegraphics[width=\wdtg]{\rt/l#1_detection}
		\end{tabular}
	}
}
\begin{figure}
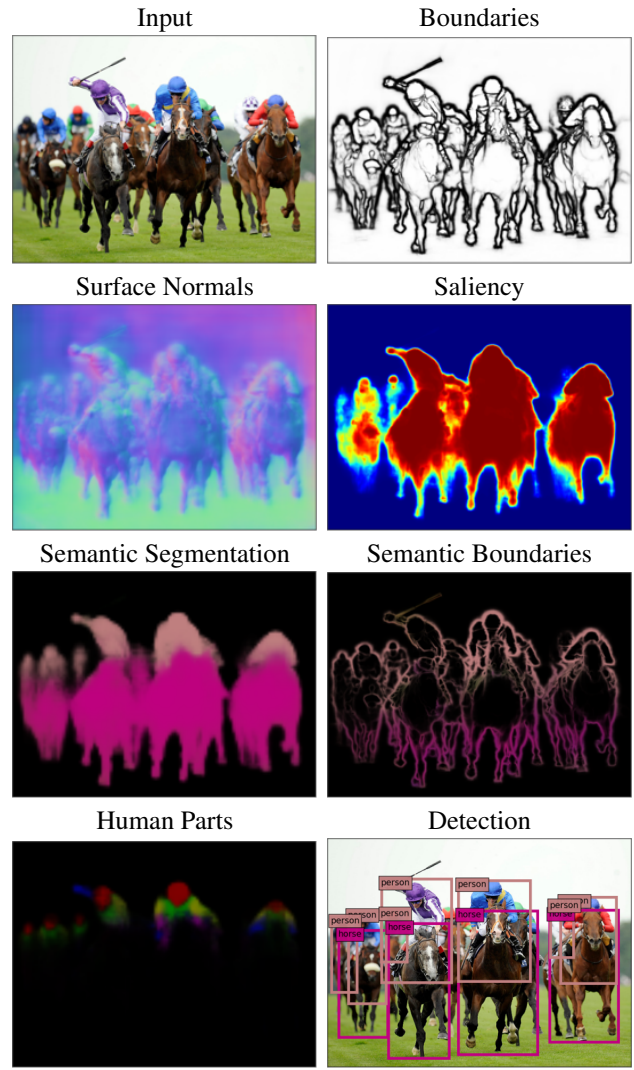

\resfigwdt{2}	
\caption{We introduce a CNN that can perform tasks spanning low-, mid- and high- level vision in a unified architecture; all results  are obtained in 0.6-0.7 seconds per frame.}
\end{figure}

\section{Introduction}


 \newcommand{\yes}{Yes}
 \newcommand{\no}{No}
 \newcommand{\few}{Partial}
 \begin{table*}
 	\begin{center}
 		\mycomment{
 			\resizebox{\textwidth}{!}{
 				\begin{tabular}{ccccccccccc}
 					& 	Imagenet \cite{ILSVRC15}&  VOC'07 \cite{pascal}	&  VOC'10 \cite{pascal}	& VOC'12 \cite{pascal} & MS-COCO \cite{mscoco} &  NYU \cite{nyu} & MSRA10K \cite{msra} & BSD \cite{MFTM01} & DTD \cite{CimpoiMKV16} & MINC \cite{BellUSB15}  \\\hline
 					Detection 	& \few & \yes 		& \yes 		& \yes 	& \yes 		& \no  & \no & \no & \no& \no \\
 					Semantic Segmentation & \no& \few & \cite{context,hariharan2011semantic} & \few &  \yes  & \yes$^{\ast}$ & \no & \no &  \no & \no\\
 					Instance Segmentation & \no & \few & \cite{context,hariharan2011semantic} & \few & \yes & \no & \no & \no &  \no & \no\\
 					Human parts & \no & \no & \cite{chen_cvpr14} & \no & \no & \no & \no & \no & \no & \no \\
 					Human landmarks & \no & \no & \cite{BourdevMM11} & \no & \yes & \no & \no & \no & \no & \no\\
 					Surface Normals & \no & \no & \no & \no & \no & \yes & \no & \no &  \no & \no\\
 					Boundaries & \no & \no & \cite{context}  & \no & \no & \no & \no & \yes&  \no & \no \\
 					Symmetry & \no & \no & \no & \no & \few,\cite{ShenZJWZB16} & \no & \no & \cite{TsogkasK12}&  \no & \no\\
 					Texture &  \no & \no & \no & \no &  \no & \no & \no & \no & \yes & \cite{CimpoiMKV16} \\
 					Materials &  \no & \no & \no & \no &  \no & \no & \no & \no & \no & \yes\\ \hline
 				\end{tabular}
 			}
 		}
 		\resizebox{\textwidth}{!}{
 			\begin{tabular}{ccccccccc}
 				& 	Imagenet \cite{ILSVRC15}&  VOC'07 \cite{pascal}	&  VOC'10 \cite{pascal}	& VOC'12 \cite{pascal} & MS-COCO \cite{mscoco} &  NYU \cite{nyu} & MSRA10K \cite{msra} & BSD \cite{MFTM01} \\\hline
 				Detection 	& \few & \yes 		& \yes 		& \yes 	& \yes 		& \no  & \no & \no  \\
 				Semantic Segmentation & \no& \few & \cite{context,hariharan2011semantic} & \few &  \yes  & \yes$^{\ast}$ & \no & \no \\
 				Instance Segmentation & \no & \few & \cite{context,hariharan2011semantic} & \few & \yes & \no & \no & \no   \\
 				Human parts & \no & \no & \cite{chen_cvpr14} & \no & \no & \no & \no & \no  \\
 				Human landmarks & \no & \no & \cite{BourdevMM11} & \no & \yes & \no & \no & \no \\
 				Surface Normals & \no & \no  & \no & \no & \no & \yes & \no & \no   \\
 				Saliency 		& \no & \no  & \no & \no & \no  & \no & \yes & \no \\
 				Boundaries 		& \no & \no & \cite{context}  & \no & \no & \no & \no & \yes  \\
 				Symmetry & \no & \no & \no & \no & \few,\cite{ShenZJWZB16} & \no & \no & \cite{TsogkasK12}\\
 			\end{tabular}
 		}
 		\caption{No single training set can accommodate all vision tasks:  several datasets contain annotations for multiple tasks, and  have  even been extended, e.g. \cite{context,chen_cvpr14,BourdevMM11}, but as the number of task grows it becomes impossible to use one dataset for all.
 			\label{datasets}}
 	\end{center}
 \end{table*}
 
 \mycomment{The representations learned by CNNs have  repeatedly been shown \cite{MahendranV15,ZeilerF14} to form pattern hierarchies,  recursively  grouping simple features  into increasingly sophisticated structures, until eventually leading to whole objects. Given that CNNs cover the whole spectrum of visual patterns, it is not surprising that they can handle both low- and high- level vision tasks.}
 
 Computer vision involves a host of tasks, such as boundary detection, semantic segmentation, surface estimation, object detection, image classification, to name a few. While 
 Convolutional Neural Networks (CNNs) have been the method of choice for text recognition for more than two decades  \cite{LeCun1998}, they have only been  recently shown to successful at handling effectively most, if not all, vision tasks. 
 
While only considering works that apply to a single, static image  we can indicatively list successes of CNNs in superresolution \cite{DongLHT16},  colorization \cite{LarssonMS16}, boundary detection \cite{shi15,ganin2014n,hed,iclr16},  symmetry detection \cite{ShenZJWZB16}, interest point detection \cite{eccv16lift}, image descriptors \cite{ZagoruykoK15,simo2015deepdesc,HanLJSB15}, 
surface normal estimation \cite{Eigen15,WangFG15,BansalRG16}, depth estimation \cite{Eigen15,LiuSL14,LiuSLR15}, intrinsic image decomposition \cite{NMY:ICCV:2015} shadow detection \cite{yago16},  texture classification \cite{CimpoiMKV16}, material classification \cite{BellUSB15}, saliency estimation \cite{ZhaoOLW15,saliencycvpr16},  semantic segmentation \cite{farabet2013learning,LongSD15,ChenPK0Y16}, region proposal generation \cite{RenHGS15,PinheiroCD15,GidarisK16}, instance segmentation \cite{hariharan2014simultaneous,PinheiroCD15,DaiHS15}, pose estimation, part segmentation, and landmark localization \cite{ZhangLLT14,ToshevS14,TsogkasSemanticPart15,ChenYWXY15,PfisterCZ15,NewellYD16,BelagiannisZ16,InsafutdinovPAA16,WeiRKS16}, as well as the large body of works around object detection and image classification e.g. \cite{KrizhevskyNIPS2013, SEZM+14, simonyan2014very, SzegedyLJSRAEVR14,girshick2014rcnn,RenHGS15,Girshick15,HeZRS15,DaiLHS16}. 
Most of these works rely on finetuning a common pretrained CNN, such as the VGG network \cite{simonyan2014very} or others \cite{KrizhevskyNIPS2013,HeZRS15}, which  indicates the broad potential of these CNNs.

 However, each of these works ends up with a task-specific CNN, and potentially a mildly different architecture. If one wanted to perform two tasks, one would need to train and test with separate networks. In our understanding a joint treatment of multiple problems can result not only in simpler, faster, and better systems, but will also be a catalyst for reaching out to other fields. One can expect that such all-in-one, ``swiss knife'' architectures will become indispensable for general AI, involving for instance   robots that will be able to recognize the scene they are in, recognize objects, navigate towards them, and manipulate them. Furthermore, having a single visual module to address a multitude of tasks will make it  possible to explore methods that  improve performance on all of them, rather than developping narrow, problem-specific techniques. Apart from simplicity and efficiency, the problem can also be motivated  by arguing that by training a network so as to accomplish multiple tasks one leaves smaller space for `blindspots' \cite{blindspots}, effectively providing a more complete specification of the network duties. 
 Finally, the particular motivation for this research has been the interest in studying the synergy between different visual tasks (e.g. the long-standing problem of combining segmentation and recognition  \cite{Keeler90,mumf94b,BottouBL97,TCYZ03,KoMa05,KTZ05,MaireYP11}), so this work can be understood as a first step in this direction. 
	
The problem of using a single network to solve multiple tasks has  been repeatedly pursued  in the context of deep learning for computer vision. In \cite{SEZM+14} a  CNN is used for joint localization  detection and classification,  \cite{Eigen15} propose a network that jointly solves surface normal estimation, depth estimation and semantic segmentation, while  \cite{GkioxariGM15}  train a system for joint detection, pose estimation and region proposal generation; \cite{ishan} study the effects of sharing information across networks trained for complementary tasks, while
more recently \cite{BilenV16} propose the introduction of inter-task connections that can improve performance through task synergy, while \cite{RanjanPC16} propose an architecture encompassing a host of face-related tasks. 

\mycomment{Addressing multiple problems through a single architecture not only  simplifies and accelerates visual processing but is also often shown to even result in performance improvements, as would be anticipated from classic multi-task learning \cite{Evgeniou04,Argyriou06}.
}

Inspired by these advances, in this work we introduce two techniques that allow us to expand the range of tasks handled by a single deep network, and thereby make it possible to train a single network for multiple, diverse tasks, without sacrificing accuracy. 

Our first contribution consists in exploring how a CNN can be trained from {\em diverse  datasets}. This problem inevitably shows up once we aim at breadth, since no single dataset currently contains ground-truth for all possible tasks. 
As shown in \reftable{datasets} high-level annotation (e.g. object positions, landmarks in PASCAL VOC \cite{pascal}) is often missing from the  datasets used for low-level tasks (e.g. BSD \cite{MFTM01}), and vice versa. If we consider for instance a network that is supposed to be predicting both human landmarks and surface normals, we have no dataset where an image comes with annotations for both tasks, but rather  disjoint datasets (NYU \cite{nyu}, and PASCAL VOC \cite{pascal}, or any other pose estimation dataset for keypoints) providing every image with annotations for only one of the two. 

\mycomment{ - and even if one such dataset existed, and we were also prepared to relabel it for any new task,  it might not have variability that would be sufficient to train any new task (consider for instance fine-grained labelling tasks). }
In order to handle this challenge we introduce in \refsec{diverse} a loss function that only relies on the ground truth available per training sample, shunning the losses of tasks for which no ground truth is available at this sample. We combine this  loss function with Stochastic Gradient Descent, and end up updating a network parameter only once we have observed a sufficient number of  training samples related to the parameter. This results in an asynchronous variant of backpropagation and allows us to train our CNN in an end-to-end manner.

\mycomment{
Furthermore,  we may have a dataset with annotations for a single task that we wish to combine with another, more thoroughly annotated dataset, and train jointly on the two. Such is the case for instance for the PASCAL VOC 2007 dataset, which comes with bounding box annotations for detection for more than 10000 images, but has minimal ground-truth for other taks - while, thanks to \cite{hariharan2011semantic,BourdevMM11,chen_cvpr14}, the PASCAL VOC 2010 dataset comes with annotations for a large set of additional tasks. }

Our second contribution aims at addressing the limitations of current hardware used for deep learning - in particular the {\em limited memory} available on modern Graphics Processing Units (GPUs).
As the number of tasks increases, the memory demand of a naively implemented back-propagation algorithm can increase linearly in the number of tasks, with a factor proportional to the memory requested by  task-specific network layers. Instead, we build on recent developments in  learning with deep architectures \cite{GruslysMDLG16,ChenXZG16} which have shown that it is possible to efficiently train a deep CNN with a  memory complexity that is sublinear in the number of layers. We develop a variant that is customized to our multi-task architecture and allows us to perform end-to-end network training for a practically unlimited number of tasks, since the memory complexity is independent of the number of tasks. 

Our current architecture has been systematically evaluated on the following tasks (a) boundary detection (b) normal estimation (c) saliency estimation (d) semantic segmentation (e) semantic part segmentation (f) semantic boundary detection 
and (g) proposal generation and object detection.
Our present system operates in 0.6-0.7 seconds per frame on a GPU and delivers  results that are competititve with the state-of-the-art for these tasks.
  

We start by specifying in \refsec{arch} the architecture of our CNN and then turn to our contributions on learning from diverse datasets  and dealing with memory constraints in  \refsec{diverse} and \refsec{memory} respectively. 

\mycomment{In normal estimation and saliency detection in particular we  obtain  results that directly compete with the most recently published techniques of \cite{Eigen15} and \cite{saliencycvpr16} respectively. In object detection we  improve the performance of a strong faster-rcnn baseline \cite{RenHGS15} from 78.7 mean Average Precision on the test set of PASCAL VOC 2007 to 80.3 by combining detection with segmentation, while the all-in-one network we finally propose recovers the original score of 78.7. 
}

\section{\ubernet architecture}
\label{arch}

\begin{figure*}
	\begin{center}
		\includegraphics[width=\linewidth]{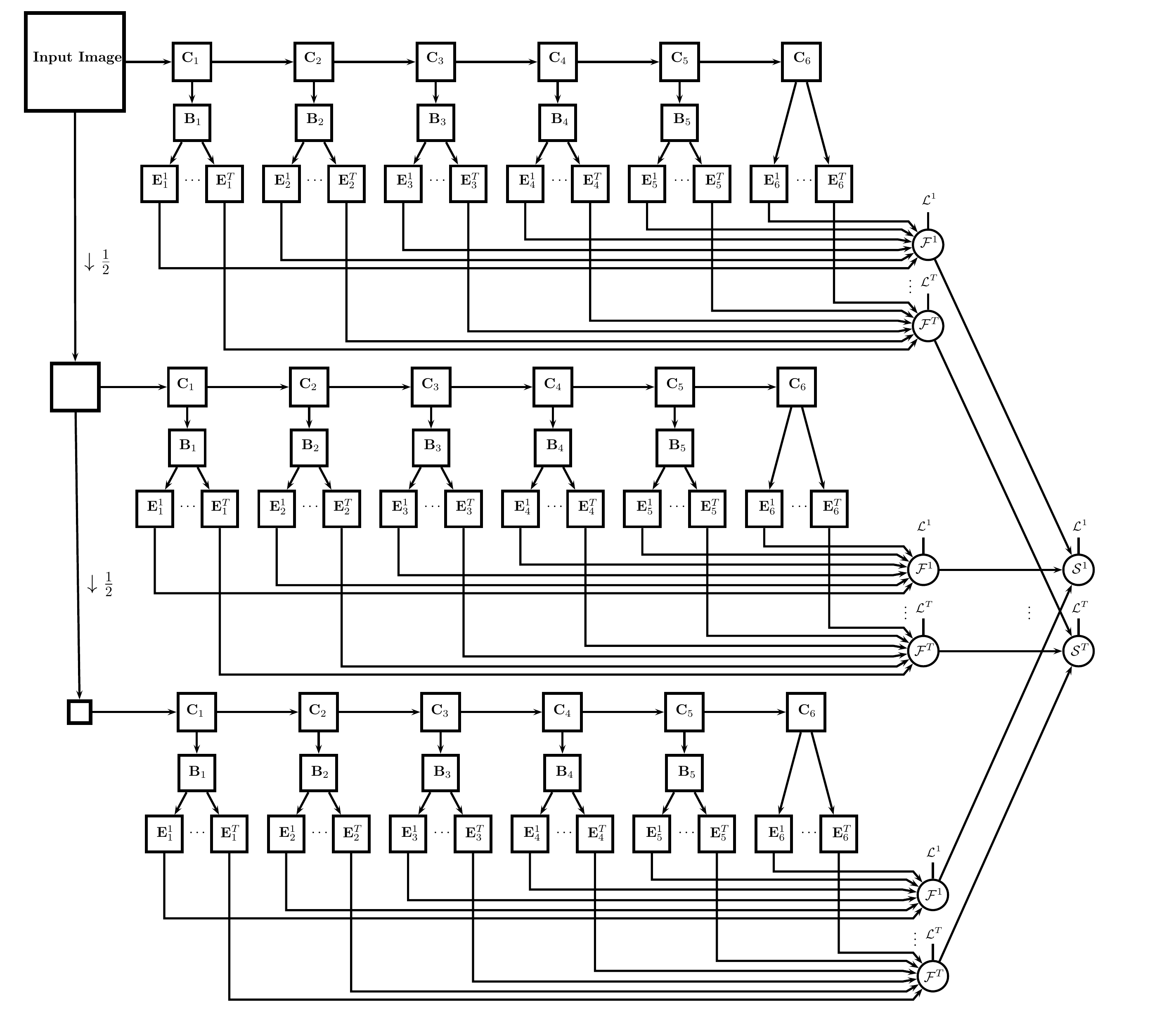}	
	\end{center}
	\caption{\ubernet architecture for jointly solving multiple labelling tasks: an image pyramid is formed by successive downsampling operations, and each image is processed by a CNN with tied weights; skip layer pooling at different network layers of the VGG network ($\mathbf{C}_i$) is combined with Batch Normalization ($\mathbf{B}_i$) to provide features that are then used to form all task-specific responses ($\mathbf{E}_i^t$); these  are combined across network layers ($\mathcal{F}^t$) and resolutions ($\mathcal{S}^t$) to form  task-specific decisions. Loss functions at the individual-scale and fused responses are used to train  task responses in a task-specific manner. For simplicity we omit interpolation, normalization and  object detection layers  - further details are provided in the text.   
		\label{fig:multi-task}}
\end{figure*}

We now introduce the  architecture of our network, shown in \reffig{fig:multi-task}. Aiming at simplicity, we introduce a minimal number of additional, task-specific  layers on top of a common  CNN trunk that is based on the VGG network. Clearly, we can always include on top additional  layers and parameters, e.g.  U-net type architectures  \cite{RonnebergerFB15,NohHH15,NewellYD16,GhiasiF16}, Dense CRF post-processing \cite{krahenbuhl2011efficient,ChenPK0Y16,crfrnn} or bilateral filter-type smoothing \cite{HarleyDK15,ChenBPMY15}, as well as more general structured prediction with  CNN-based pairwise terms \cite{Adelaide,ChandraK16} - but we leave this for future work.



 The starting point is that of using a standard  `fully' convolutional network  \cite{LeCun1998,SEZM+14,PapandreouKS15,LongSD15}, namely a CNN that provides   a field of decision variables, rather than a single classificiation at its output; this can be used to accomplish any dense labelling or regression task, such as boundary detection, normal estimation, or semantic segmentation. We now describe our modifications to this basic architecture.

{\textbf{Skip layers}}: one first deviation from the most standard architecture is that as in \cite{SEZM+14,hariharan2014hypercolumns,hed,iclr16} we use skip layers that combine the top-layer neurons with the activations of intermediate neurons to form the network output. Tasks such as boundary detection clearly profit from the smaller degree of spatial abstraction of lower-level neurons \cite{hed}, while even for high-level tasks, such as semantic segmentation, it has been shown \cite{hariharan2014hypercolumns,LongSD15} that  skip layers can improve performance. In particular  we pool features from layers  $\mathrm{conv1\_2, conv2\_2, conv3\_3, conv4\_3, conv5\_3, fc7}$ of the VGG-16 network, which show up as $\mathbf{C}_1,\ldots,\mathbf{C}_6$ in \reffig{fig:multi-task}. 

{\textbf{Skip-layer normalization:}} Modifying slightly \cite{parsenet,bell16ion}, we  use batch normalization \cite{batchnorm} prior to forming the inner product with intermediate layers; this alleviates the need for very low learning rates, which was the case in \cite{hed,iclr16}. One exception is the last layer ($\mathrm{fc7}$ in VGG, $\mathbf{C}_6$ in \reffig{fig:multi-task}) which has already been trained so as to be an appropriate argument to a linear classifier ($\mathrm{fc8}$), and therefore seemed  to be doing better without normalization.


\newcommand{\neuron}{\mathbf{f}}
\newcommand{\scoret}{\mathbf{s}}
\newcommand{\weight}{\mathbf{w}}
\newcommand{\is}{}

{\textbf{Cumulative task-specific operations:}}
Scaling up to many tasks requires  keeping the task-specific memory  and computation budget low, and we therefore choose as in  \cite{hed,iclr16} to process the outputs of the skip-pooling with task-specific layers that perform linear operations.   In particular if we denote by
 $\neuron_1\is,\ldots,\neuron_6\is$ the  neuron activations at layers $1$ up to $6$ that are used to obtain the score at a given image position 
  the output for task $t$ is a linear function:
 \begin{gather}
 \scoret_t = \weight^T_t \neuron  = \sum_{k=1}^{6} \weight^T_{t,k} \neuron_k \\
 \mathrm{with}\quad \neuron =
 \left[\neuron_1 |	  \hdots | \neuron_6 \right]^T, \weight_t = \left[
 \weight_{t,1} |
 \hdots | \weight_{t,6}
 \right]^T.
 	 \end{gather}	 
 	Rather that explicitly allocating memory for the vector $\neuron$ formed by contactenating the intermediate activations and then forming the matrix product,
 	we instead compute the intermediate results per layer, $\weight^T_{t,k} \neuron_k$ and then add them up. This  yields the same result, but acts like a low-memory  online accumulation of scores across skip layers. 
 	
 	{\textbf{Fusion layers:}}
 	For the fusion layers, denoted by circular connections in \reffig{fig:multi-task} we observe that instead of simply adding the scores (sum-fusion), one can  accelerate training by concatenating {{the score maps}} and learning a linear function that operates on top of the concatenated score maps - as originally done in \cite{hed}. This scheme is clearly still learning in the end a linear function with the same number of free parameters, but this decomposition, which can be intuitively understood as some form of preconditioning, seems to be more effective. When also back-propagating on the intermediate layers, this typically also results in better performance.  We note that for simplicity we  assume   correspondence across layer positions; these are handled in our network by appropriate interpolation layers which in our diagram are understood to be included in the circular nodes. 
 	
\textbf{{Atrous convolution:}}
We also use {{convolution with holes}} (\`a trous) \cite{PapandreouKS15,ChenPK0Y16} which allows us to control the spatial resolution of the output layer. In particular we use \`a trous convolution to obtain an output stride of 8 (rather than 16), which gives us a moderate boost in tasks such as boundary detection or semantic segmentation. 
 	

\textbf{{Multi-resolution CNN:}} as in \cite{jacobs14,PapandreouKS15,iclr16,ChenYWXY15}, rather than processing an image at a single resolution, we form an image pyramid and pass  scaled versions of the same image through CNNs with shared weights. This allows us to deal with the scale variability of image patterns. Even though in \cite{ChenYWXY15} a max-fusion scheme is shown to yield higher accuracy than sum-fusion, in our understanding this is particular to the case of semantic segmentation, where a large score at any scale suffices to assign an object label to a pixel. This may not be the case for boundaries where the score should be determined by the accumulation of evidence from multiple scales \cite{witkin} or for normals, where maximization over scales of the normal vector entries does not make any sense. 
We therefore use a concatenation of the scores followed by a linear operation, as in the case of fusing the skip-layers described above, and leave the exploration of  scale-aware processing \cite{ChenYWXY15} for the future.

This multi-resolution  processing is incorporated in the network definition, and as such can be accounted for during end-to-end training.  
In this pyramid the highest resolution image is set similar to \cite{Girshick15} so that the smallest image dimension is 621 pixels and the largest dimension does not exceed 921 (the exact numbers are so that dimensions are of the form $32k +1$, as requested by \cite{iclr16,ChenYWXY15}).

As in \cite{iclr16,ChenYWXY15} we use loss layers both at the outputs of the individual scales and  the final responses, amounting to a mild form of deep supervision network (DSN)  training \cite{hed}.

{\textbf{Task-specific deviations:}}
All of these choices have been separately validated on individual tasks, and then integrated in the common architecture shown in \reffig{fig:multi-task}. 
There are still however some task-specific deviations.

 One  exception to the uniform architecture  outlined above is for detection, where we follow the work of \cite{RenHGS15} and learn a convolutional region proposal network,  followed by a fully-connected subnetwork that classifies the region proposals into one of  21 labels (20 classes and background). Recent advances however \cite{DaiLHS16,ssd}  may make this exception unnecessary.
 

Furthermore, the output of each of the task-specific streams is penalized by a loss function that is adapted to the task at hand. 
 For region labelling tasks (semantic segmentation, human parts, saliency) and object detection we use the softmax loss function, as is common in all recent works on semantic segmentation \cite{LongSD15,Chen2015iclr} and object detection \cite{Girshick15}. For regression tasks (normal estimation, bounding box regression)
 we use the smooth $\ell_1$ loss \cite{Girshick15}. For normal estimation we apply an $\ell_2$ normalization prior to penalizing with the $\ell_1$ loss, since surface normals are unit-norm vectors. 
 For tasks where we want to estimate thin structures (boundaries, semantic boundaries) we use the MIL-based loss function introduced in \cite{iclr16} in order to  accommodate imprecision in the placement of the boundary annotations. For these two tasks we also have a  class imbalance problem, with many more negatives than positives; we mitigate this by using a weighted cross-entropy loss, as in \cite{hed}, where we  attribute a weight of $0.9$ to positives and $0.1$ to negatives. 
 
Furthermore, for  low-level tasks such as boundary detection, normal estimation and saliency estimation, as well as semantic boundary detection we set the spatial resolution of the scores to be  equal to that of the image - this allows us to train with a loss function that has a higher degree of spatial accuracy and allows us to accurately localize small structures. For region labelling tasks such as semantic segmentation, or human part segmentation we realized that we do not really need this level of spatial accuracy, and instead train score maps that employ a lower spatial resolution, using a downsampling factor of 8 with respect to the original image dimensions. This results in a 64-fold reduction of  the task-specific computation and memory demands. 

%
 
\mycomment{
Most recent works aim at solving problems that are at a common representation, e.g. mid-level (surface, depth, region labels in \cite{Eigen14}, or pose and category labels in \cite{Gkioxari15}. In this work we consider a combination of two tasks where the desired outputs come in different formats (regions for detection, pixel labellings for segmentation) and explore the interplay between them. 
}

\mycomment{
A complementary line of progress has been the fusion of segmentation and detection; for this we use a `two-headed' network, where one head is fully-convolutional until the end, and addresses the semantic segmentation task, as in \cite{iclr16,ChenPKMY14,long_shelhamer_fcn}, while the other relies on region proposals to process a shortlist of interesting positions, as in \cite{RenHGS15}.
Our first experiments in this direction  demonstrated that the two tasks can benefit from each other by being solved jointly in two ways; firstly, just training a network that performs two tasks turned out to improve performance - and secondly by interleaving the segmentation and detection tasks we have been able to provide information from one task to the other. Our current results indicate that through this procedure we can improve detection performance from 78.7 to 80.3 on the PASCAL VOC test set, as shown in \reftable{table:results_ap}.
}








 \section{Multi-Task Training using Diverse Datasets}
\label{datasets}
\label{diverse}
 \newcommand{\viss}[1]{\mathbf{V}}
 \newcommand{\vis}[1]{\mathbf{v}_{#1}}
 \newcommand{\prd}[2]{\mathbf{f}^{#2}_{#1}}
 \newcommand{\gt}[2]{\mathbf{y}^{#2}_{#1}}
 \newcommand{\raw}[2]{\mathbf{I}^{#1}_{#2}}
 \newcommand{\prmm}[1]{\mathbf{W}_{#1}}
 \newcommand{\prm}[1]{\mathbf{w}_{#1}}
 \newcommand{\cnnw}[1]{\mathbf{w}^{#1}}
 \newcommand{\prms}[1]{\mathbf{\ww}_{#1}}
 \newcommand{\reg}[1]{\mathcal{R}(#1)}
 \newcommand{\task}{t}
 \newcommand{\losst}[3]{L_{#1}\left(#2,#3\right)}
  \newcommand{\losstwo}[2]{L_{#1}\left(#2\right)}
  \newcommand{\lossd}[3]{L'_{#1}\left(#2,#3\right)}
\newcommand{\lossdd}[4]{\nabla_{#4}L_{#1}\left(#2,#3\right)}
 \newcommand{\mydelta}[2]{\delta_{#1,#2}}
 \newcommand{\cnn}{0}
 Having described our network's architecture, we now turn to  parameter estimation. 
 Our objective is to train in an end-to-end manner both the VGG-based CNN trunk that delivers features to all tasks, and the weights of the task-specific layers.

\mycomment{
 k, since the high-level tasks are trying to optimize over a network that is a `moving target' - while freezing the network weights to some values determined from a high level task always results in worse performance in the low-level tasks. 
We have adapted  backpropagation training to asynchronously update network parameters, so that parameters specific to a task are updated only once sufficient annotated training samples have been observed. The minibatch construction has been accordingly modified so that a proper blend of the different datasets is contained in every minibatch. 
}


As described in the introduction, the main challenge that we face is the diversity of the tasks that we wish to cover. 
\mycomment{
For the standard multi-task training problem one can simply consider the sum of the per-task losses as the training objective, and minimize it  through back-propagation, as done e.g. in \cite{Eigen15,GkioxariGM15,DaiHS15}.
This however can be become challenging if one wants to jointly tackle problems for which no common dataset exists.}
 %
%
%
In order to handle the diversity of the available datasets one needs to handle  missing ground truth data during training. Certain recent works such as \cite{papa15,OquabBLS15,dai2015boxsup} manage to impute missing data in an EM-type approach, by exploiting domain-specific knowledge - e.g. by requesting that a fixed percentage of the pixels contained in the  bounding box of an object obtain the same label as the object. This however may not be possible for arbitrary tasks, e.g. normal estimation. 

Instead, we propose to adapt the loss function to the information that we have per sample, and set to zero the loss of tasks for which we have no ground-truth. While the idea is straightforward, as we describe below some care needs to be taken when optimizing the resulting loss with backpropagation, so as to ensure that the (stochastic) estimates of the parameter gradients accumulate evidence from a sufficient number of training samples.  


Our training objective is expressed as the sum of per-task losses, and regularization terms applied to the parameters of task-specific, as well as shared layers:
\begin{gather}
\Loss(\prm{\cnn,\task_1,\ldots,\task_T}) = \reg{\prm{\cnn}} + \sum_{t=1}^T  \gamma_{\task_k} \left(\reg{\prm{\task}}  +  \losstwo{\task}{\prm{\cnn},\prm{\task}}\right).
\label{eq:objective}
\end{gather}
In \refeq{eq:objective} we use $\task$ to index tasks; $\prm{\cnn}$ denotes the weights of the common CNN trunk,  and $\prm{{\task}}$ are task-specific weights; $\gamma_t$ is an hyperparameter that determines the relative importance of task $t$,
 $\reg{\prm{\mathrm{\ast}}} =  \frac{\lambda}{2}\|\prm{\mathrm{\ast}}\|^2$ is an $\ell_2$ regularization on the relevant network weights, and $\losstwo{\task}{\prm{\cnn},\prm{\task}}$ is the task-specific loss function.


This task-specific loss  is written as follows:
\begin{gather}
\losstwo{\task}{\prm{\cnn},\prm{\task}} = \frac{1}{N}\sum_{i=1}^{N}  \mydelta{\task}{i} \losst{\task}{\prd{\task}{i}(\prm{\cnn},\prm{\task})}{\gt{\task}{i}},
\end{gather}
where we use $i$ to index training samples, denote by $\prd{\task}{i}$,
$\gt{\task}{i}$  the task-specific network prediction and ground truth at the $i$-th example respectively, by $\prm{i}$ the task-specific network parameters, and by $\mydelta{\task}{i} \in \{0,1\}$ we indicate whether  example $i$ comes with ground-truth for  task $\task$.
If $\mydelta{\task}{i}=0$ we can set $\gt{\task}{i}$ to an arbitrary value, without affecting the loss - i.e. we do not need to impute the ground truth. 

We now turn to accounting for  the interplay  between the term $\mydelta{\task}{i}$ and the minimization of \refeq{eq:objective}
through Stochastic Gradient Descent (SGD). Since we want to train our network in an end-to-end manner for all tasks, we consider as our training set the union of different training sets, each of them containing pairs of images and ground-truth for distinct tasks. Images from this set are sampled uniformly at random; as is common practice, we  use multiple epochs and within each epoch sample without replacement. 



\newcommand{\bi}{i}
\newcommand{\bs}{B}
\newcommand{\mb}{\mathcal{B}}
\newcommand{\tast}{p}

 \newcommand{\grad}[1]{{\mathrm{\mathbf{d}}#1}}
 \newcommand{\cnt}[1]{\mathrm{\mathbf{c}}_{#1}}
 \newcommand{\setto}{\leftarrow}
 \begin{table}[!t]
 	\hrule
 	\vspace{3pt}
 	Synchronous SGD - backprop
 	\hrule
 	\begin{algorithmic}
 		\FOR{$m = 1$ to $M$}   
 		\STATE \COMMENT{construct minibatch}
 		\STATE  $\mb \setto \{\bi_1,\ldots,\bi_{\bs}\}$ with $\bi_i \sim U[1,N]$ 
 		\STATE \COMMENT{initialize gradient accumulators}
 		\STATE $\grad{\prm{\cnn}} \setto 0,\grad{\prm{1}} \setto 0,\ldots,\grad{\prm{T}} \setto 0$  
 		\FOR{$\bi \in \mb$}
 		\STATE \COMMENT{cnn gradients}
 		\STATE $ \grad{\prm{\cnn}} \setto \grad{\prm{\cnn}} + \sum_{\task} \mydelta{\task}{\bi}\gamma_{\task}\lossdd{\task}{\prd{\task}{\bi}(\prm{\cnn},\prm{\task})}{\gt{\task}{\bi}}{\prm{\cnn}}$
 		\STATE \COMMENT{Task gradients, $t=1,\ldots,T$}
 		\STATE $ \grad{\prm{\task}} \setto \grad{\prm{\task}} + \mydelta{\task}{\bi}\gamma_{\task}\lossdd{\task}{\prd{\task}{\bi}(\prm{\cnn},\prm{\task})}{\gt{\task}{\bi}}{\prm{\cnn}}$ 
 		\ENDFOR
 		\FOR{$\tast \in \{\cnn,1,\ldots,T\}$}
 		\STATE $\prm{\tast} \setto \prm{\tast} - \epsilon\left( \lambda  \prm{\tast}   + \frac{1}{\bs}\grad{\prm{\tast}} \right)$
 		\ENDFOR
 		\ENDFOR
 		\vspace{.2cm}
 		\hrule
 	\end{algorithmic}
 	\caption{Pseudocode for the standard, synchronous stochastic gradient descent algorithm for back-propagation training. We update all parameters at the same time, after observing a fixed number of samples. \label{sync}}
 \end{table}
 
 \newcommand{\bstask}{\bs_{p}}
 \begin{table}
 	\hrule
 	\vspace{3pt}
 	Asynchronous SGD - backprop
 	\hrule
 	\begin{algorithmic}
 		\STATE \COMMENT{initialize gradient accumulators}
 		\STATE $\grad{\prm{\cnn}} \setto 0 , \grad{\prm{1}} \setto 0,\ldots,\grad{\prm{T}} \setto 0$  
 		\STATE \COMMENT{initialize  counters}
 		\STATE $\cnt{\cnn} \setto 0,\cnt{1} \setto 0,\ldots,\cnt{T} \setto 0$  
 		\FOR{$m = 1$ to $M\cdot \bs$}   
 		\STATE  Sample $\bi \sim U[1,N]$ 
 		\STATE \COMMENT{cnn gradients \& counter: always updated}
 		\STATE $\cnt{\cnn} \setto  \cnt{\cnn} + 1$
 		\STATE $\grad{\prm{\cnn}} \setto \grad{\prm{\cnn}} + \sum_{\task} \mydelta{\task}{\bi}\gamma_{\task}\lossdd{\task}{\prd{\task}{\bi}(\prm{\cnn},\prm{\task})}{\gt{\task}{\bi}}{\prm{\cnn}}$
 		
 		\FOR{$\task \in \{1,\ldots, T\}$}
 		\IF{$\mydelta{\task}{\bi}=1$}
 		\STATE \COMMENT{update accumulator and counter for  task $\task$ if the current sample is relevant}
 		\STATE $\cnt{\task} \setto  \cnt{\task} +  1$
 		\STATE $ \grad{\prm{\task}} \setto \grad{\prm{\task}} + \gamma_{\task}\lossdd{\task}{\prd{\task}{\bi}(\prm{\cnn},\prm{\task})}{\gt{\task}{\bi}}{\prm{\task}}$ 
 		\ENDIF
 		\ENDFOR	
 		\FOR{$\tast \in \{\cnn,1,\ldots,T\}$}
 		
 		\IF {$\cnt{\tast}=\bstask$}
 		\STATE \COMMENT{update parameters if we have seen enough}
 		\STATE $\prm{\tast} \setto \prm{\tast} - \epsilon\left( \lambda  \prm{\tast}   + \frac{1}{\bstask}\grad{\prm{\tast}} \right)$
 		\STATE $\cnt{\tast}\setto 0 $, $\grad{\prm{\tast}} \setto 0$
 		\ENDIF
 		\ENDFOR
 		\ENDFOR
 		\vspace{.2cm}
 		\hrule
 	\end{algorithmic}
 	\caption{Pseudocode for our asynchronous stochastic gradient descent algorithm for back-propagation training. We update a task-specific parameter only after observing  sufficient many training samples that pertain to the task. \label{async}}
 \end{table}

Considering that we use a  minibatch $\mb$  of size $\bs$, plain SGD for task $k$ would lead to the following update rules:  
\begin{eqnarray}
\!\!\!\!\prm{\tast}' \!\!&=&\!\! \prm{\tast} - \epsilon (\lambda \prm{\tast} + \mathbf{dw}_{\tast}), \quad \tast \in \{\cnn,1,\ldots,T\}\\
\!\!\!\!\mathbf{dw}_{\cnn} \!\!&=& \!\!  \frac{1}{\bs}\sum_{\bi \in \mb}\sum_{\task=1}^T \gamma_{\task}  \mydelta{\task}{\bi} \lossdd{\task}{\prd{\task}{\bi}(\prm{\cnn},\prm{\task})}{\gt{\task}{\bi}}{\prm{\cnn}},  
\\
\!\!\!\!\mathbf{dw}_{\task} \!\!&=&\!\!  \frac{1}{\bs}\sum_{\bi \in \mb} \gamma_{\task}\mydelta{\task}{\bi}\lossdd{\task}{\prd{\task}{\bi}(\prm{\cnn},\prm{\task})}{\gt{\task}{\bi}}{\prm{\mathrm{\task}}}, 
\end{eqnarray}
where the weight decay term results from $\ell_2$ regularization and $\lossdd{\task}{\hat{y}}{y}{\prm{\ast}}$
denotes the gradient of the loss for task $\task$  with respect to the parameter vector ${\prm{\ast}}$. The difference between the two update terms is that the parameters of the common  trunk, $\prm{\cnn}$ are affecting all tasks, and as such accumulate the gradients over all tasks, while task-specific parameters $\prm{\task}$ are only affected by the subset of images for which $\mydelta{\task}{\bi}=1$. 


We observe that this comes with an important flaw: if we have a small batch size, the update rule for $\prm{\task}$ may use a too noisy gradient if  $\sum_{\bi \in  \mb} \mydelta{\task}{\bi}$ happens to be small  -  it may even be that no task-related samples happen to be in the present minibatch. We have empirically observed that this can often lead to erratic behaviour, which we originally handled  by increasing the minibatch size to quite large numbers (50 images or more, as opposed to 10 or 20). Even though this mitigates the problem partially, firstly it is highly inefficient timewise, and will also not scale up to solving say 10 or 20 tasks simultaneously. 


Instead of this brute-force approach we propose a modified variant of backpropagation that more naturally handles the problem
by updating the parameters of a task only once sufficiently many relevant images have been observed. 
Pseudocode presenting this scheme, in contrast to the standard SGD scheme is provided in \reftable{async}. 

In particular we no longer have `a minibatch', but rather treat images in a streaming mode, keeping one counter per task - as can be seen, rather than $M$ outer loops, which is the number of minibatches, we use $M\cdot\bs$ outer loops, equalling the number of images treated by the original scheme. 

 Whenever we process a training sample that contains ground truth for a task, we increment the task counter, and add the current gradient to a cumulative gradient sum. Once the task counter exceeds a threshold we update the task parameters and then reset the  counter and cumulative gradient to zero. 
 Clearly, the common CNN parameters are updated regularily, since their counter is  incremented for every single training image. This is however not the case for other tasks which may not be affected by a subset of training images. 
 
 This results in an asynchronous variant of backpropagation, in the sense that any parameter can be updated at a time instance that is independent of the others. We note that apart from implementing the necessary book-keeping, this scheme requires no additional memory or computation. It is also clear that the `asynchronous' term relates to the manner in which parameters for different tasks are updated, rather than the computation itself, which in our implementation is single-node.

We also note that according to the pseudocode, we allow ourselves to use different `effective batchsizes', $\bstask$, which we have observed to be useful for training. In particular, for detection tasks it is reported in \cite{Girshick15} that a batchsize of two suffices, while for dense labelling tasks such as  semantic segmentation a batchsize of 10, 20 or even 30 is often used  \cite{hed,ChenPK0Y16}.
In our training we use an effective batchsize  $\bstask$ of 2 for detection, $10$ for all other task-specific parameters, and $30$ for the shared CNN features, $\prm{\cnn}$. The reasoning behind  using this larger batch size for the shared CNN features is that we want their updates to absorm information from a larger number of images, containing multiple tasks, so that the task-specific idiosyncracies will cancel out. In this way it becomes more likely that 
the average gradient will serve all tasks and  we avoid having the `moving target' problem, where every task quickly changes the shared representation of the other tasks,  making optimization harder. 

One subtle difference is that in synchronous SGD the stochastic estimate of the gradient used in the update equals:
\ba
\mathbf{g_{\task}^s} = \gamma_{\task}\frac{1}{\bs}\sum_{i \in \mb} \mydelta{\task}{\bi} \lossdd{\task}{\prd{\task}{\bi}(\prm{\cnn},\prm{\task})}{\gt{\task}{\bi}}{\prm{\mathrm{\task}}}\ea
while for the asynchronous case it will equal:
\ba
\mathbf{g_{\task}^a} =
\gamma_{\task}\frac{1}{\bstask}\sum_{i \in I_{\bstask}}  \lossdd{\task}{\prd{\task}{\bi}(\prm{\cnn},\prm{\task})}{\gt{\task}{\bi}}{\prm{\mathrm{\task}}},\ea
where $I_{\bstask} =\{i_{\task,1},\ldots,i_{\task,\bstask}\}$ indicates a subsequence of $\bstask$ samples which contain ground-truth for task $\task$. We realize that the first estimate can be expected to have a typically smaller magnitude that the second one, since several of the terms being averaged will equal zero. This implies that we have somehow modified the original cost function, since the stochastic gradient estimates do not match.
However this effect can be absorbed in the (empirically set) hyperparameters $\gamma_t$ so that the two estimates will have the same expected magnitude, so we can consider the two algorithms to be optimizing the same quantity.  








\newcommand{\nlcnn}{L_{C}}
\newcommand{\nltask}{L_{T}}
\newcommand{\nbytes}{N}

\newcommand{\wdf}{.5\textwidth}
\begin{figure}[!h]
	\begin{center}
		\includegraphics[width=\wdf]{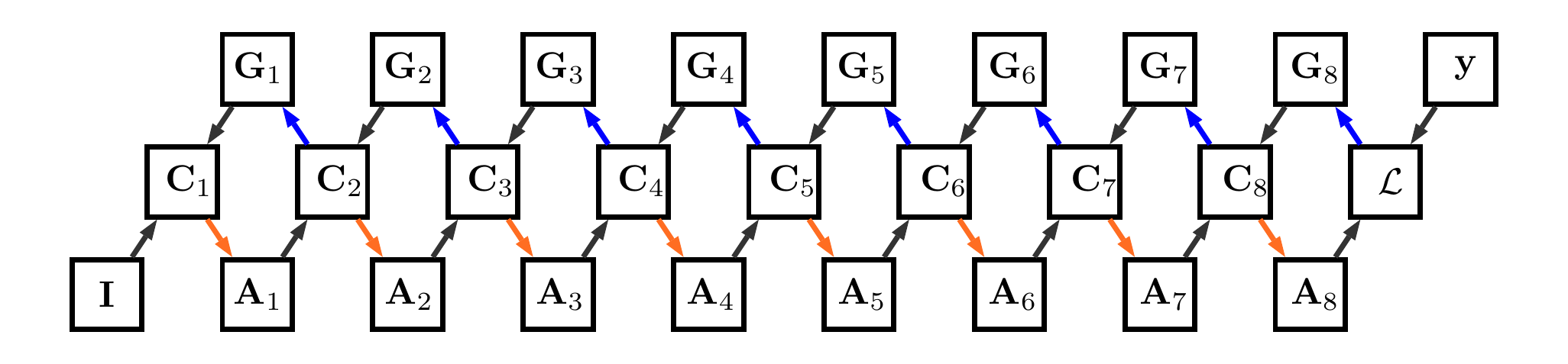}
	\end{center}
	\caption{Vanilla backpropagation for a single task; memory lookup operations are indicated by black arrows, storage operations are indicated by orange and blue arrows for the forward and backward pass respectively.
		During the forward pass each layer stores its activation signals in the bottom boxes. During the backward pass these activation signals are  combined with the gradient signals (top boxes) that are  computed recursively, starting from the loss layer.  \label{fig:vanillasingle}}
\end{figure}

\begin{figure}
	\begin{center}
		\subfloat[Low-memory forward pass]{\includegraphics[width=\wdf]{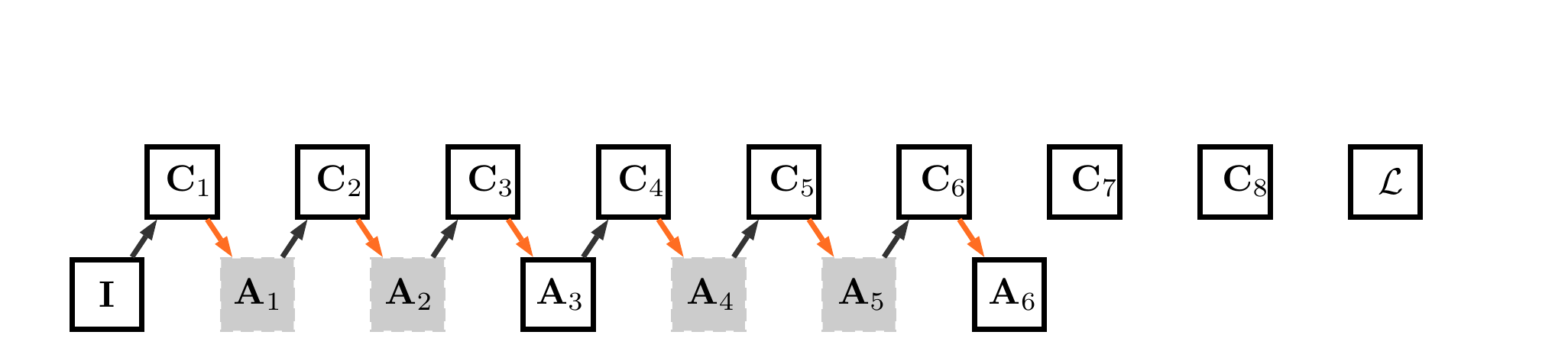}}\\
		\subfloat[Low-memory backpropagation (7-9)]{\includegraphics[width=\wdf]{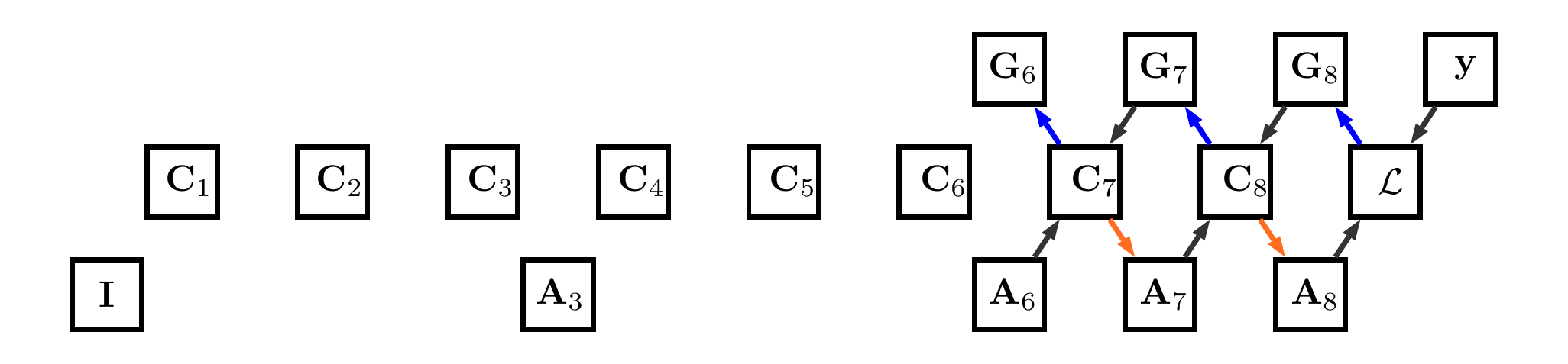}}\\
		\subfloat[Low-memory backpropagation (4-6)]{\includegraphics[width=\wdf]{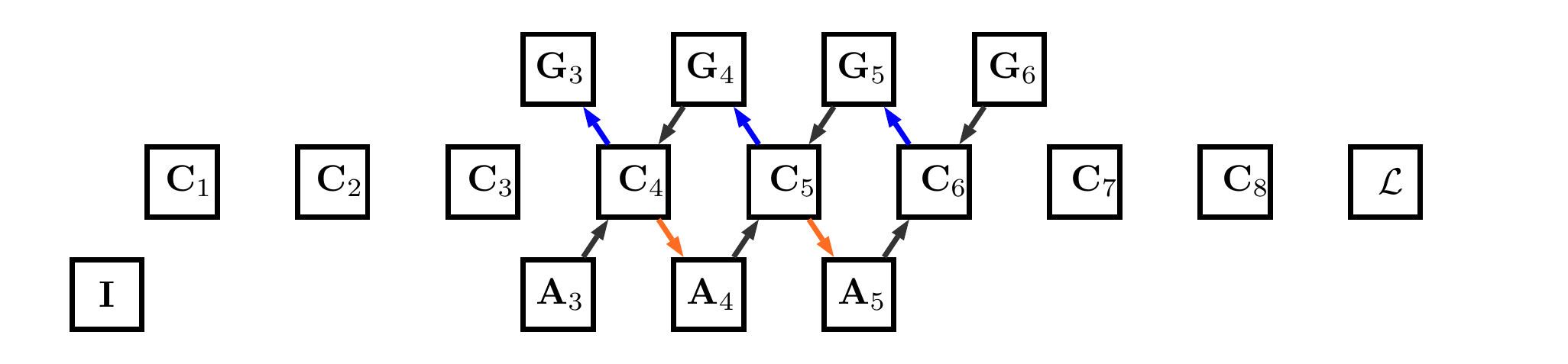}}\\
		\subfloat[Low-memory backpropagation (1-3)]{\includegraphics[width=\wdf]{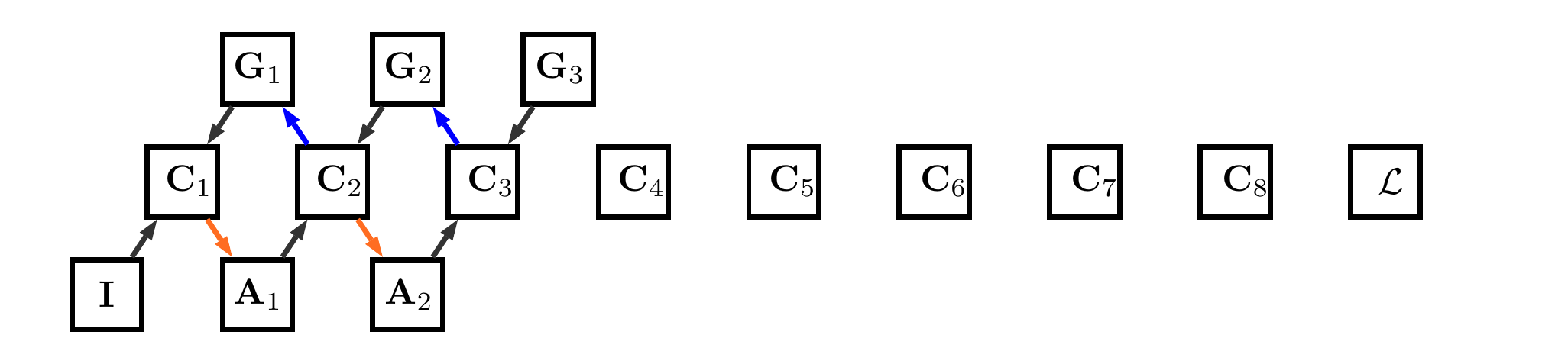}}
	\end{center}
	\caption{Low-memory backpropagation for a single task (same color code as in \reffig{fig:vanillasingle}). We first store a subset of activations in memory, that then serve as  `anchor' points for running backpropagation on smaller networks. This reduces the number of layer activations/gradients that are simultaneously  stored in memory. \label{fig:memorysingle}}
\end{figure}

\begin{figure}
	\includegraphics[width=\wdf]{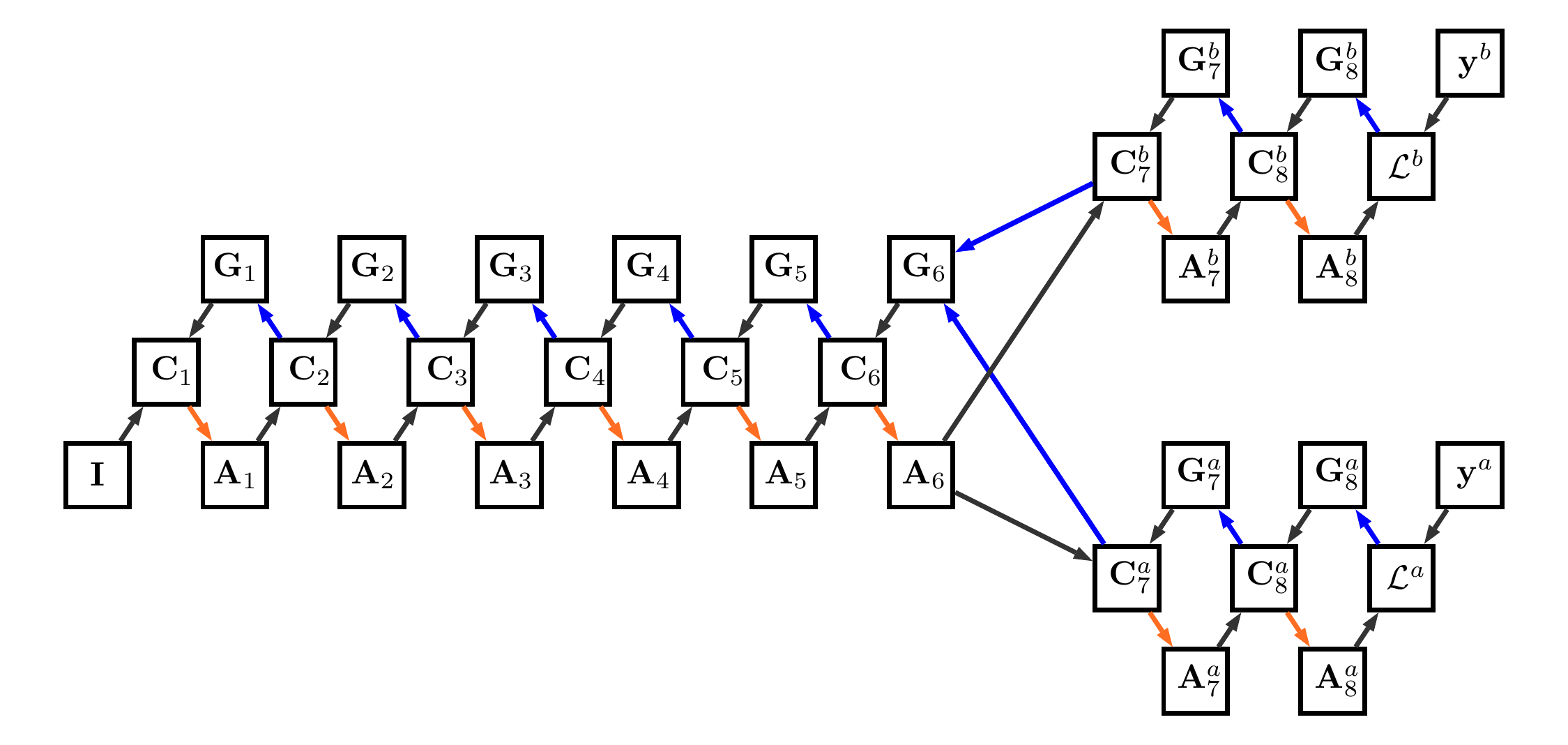}
	\caption{Vanilla backpropagation for multi-task training: a naive implementation has a memory complexity  $2 \nbytes (\nlcnn   +  T \nltask )$, where here $\nlcnn=6$ is the depth of the common CNN trunk, $\nltask=3$ is the depth of the task-specific branches and $T=2$ is the number of tasks.\label{fig:vanillamtask}}
\end{figure}

\newcommand{\trm}{ .4cm }
\begin{figure}
	\subfloat[Low-memory forward pass]{\includegraphics[trim={0 1.1cm 0 1.1cm }, width=\wdf]{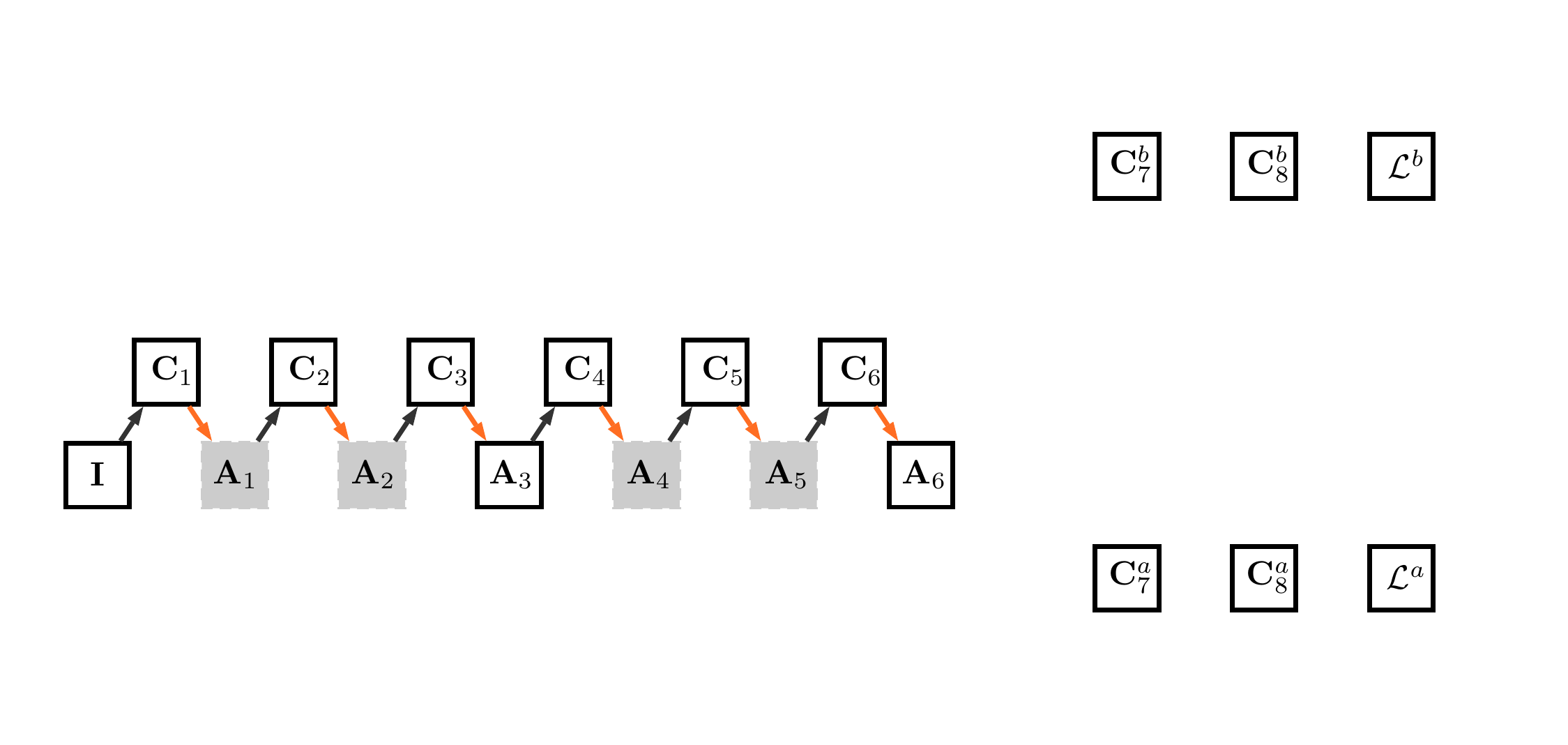}}\\
	\subfloat[Low-memory backpropagation - task a]{\includegraphics[trim={0 0 0 1.1cm },width=\wdf]{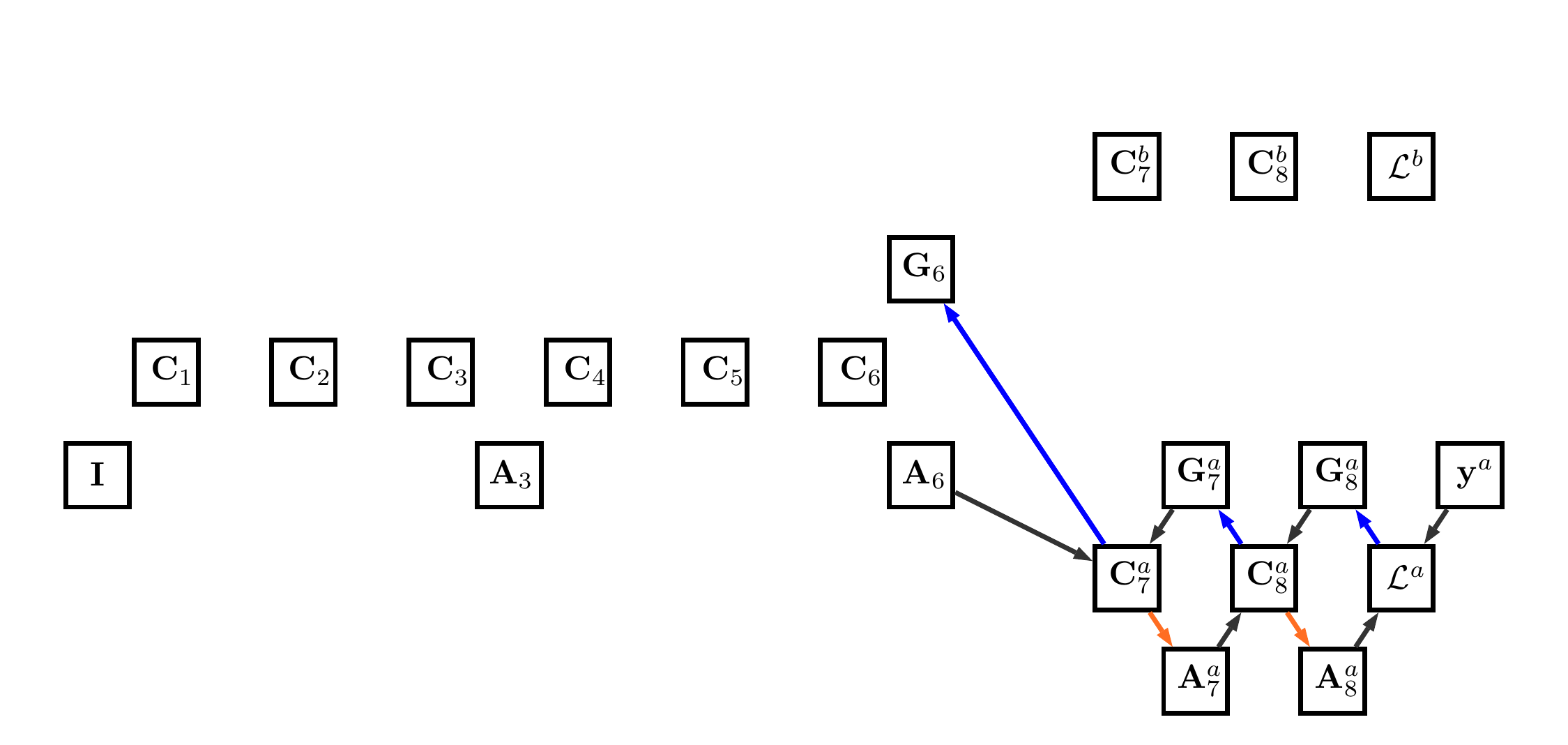}}\\
	\subfloat[Low-memory backpropagation - task b]{\includegraphics[trim={0 1.1cm  0 0},width=\wdf]{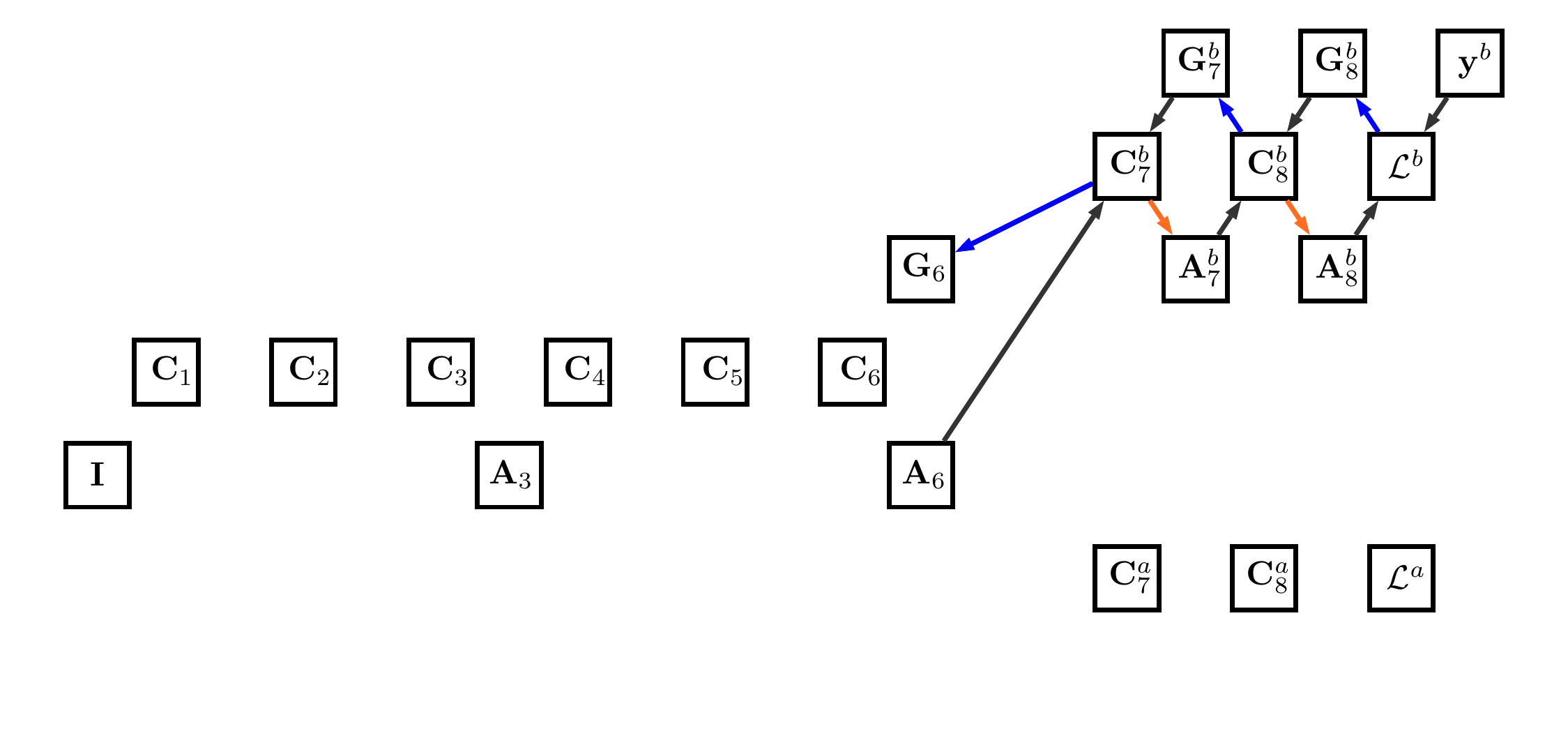}}\\
	\subfloat[Low-memory backpropagation (4-6) ]{\includegraphics[trim={0 1.1cm  0 1.1cm },width=\wdf]{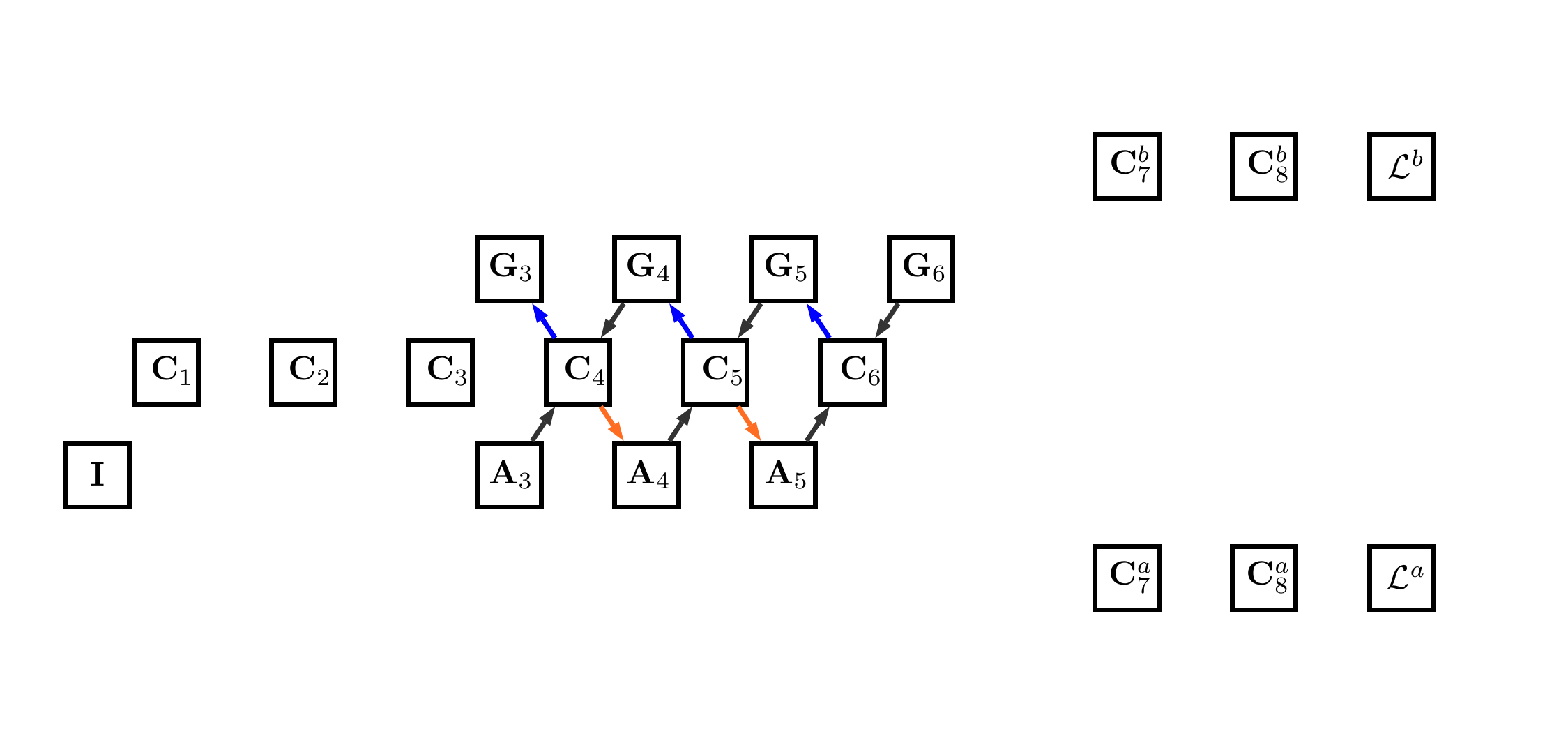}}\\
	\subfloat[Low-memory backpropagation (1-3)]{\includegraphics[trim={0 1.1cm  0 1.1cm },width=\wdf]{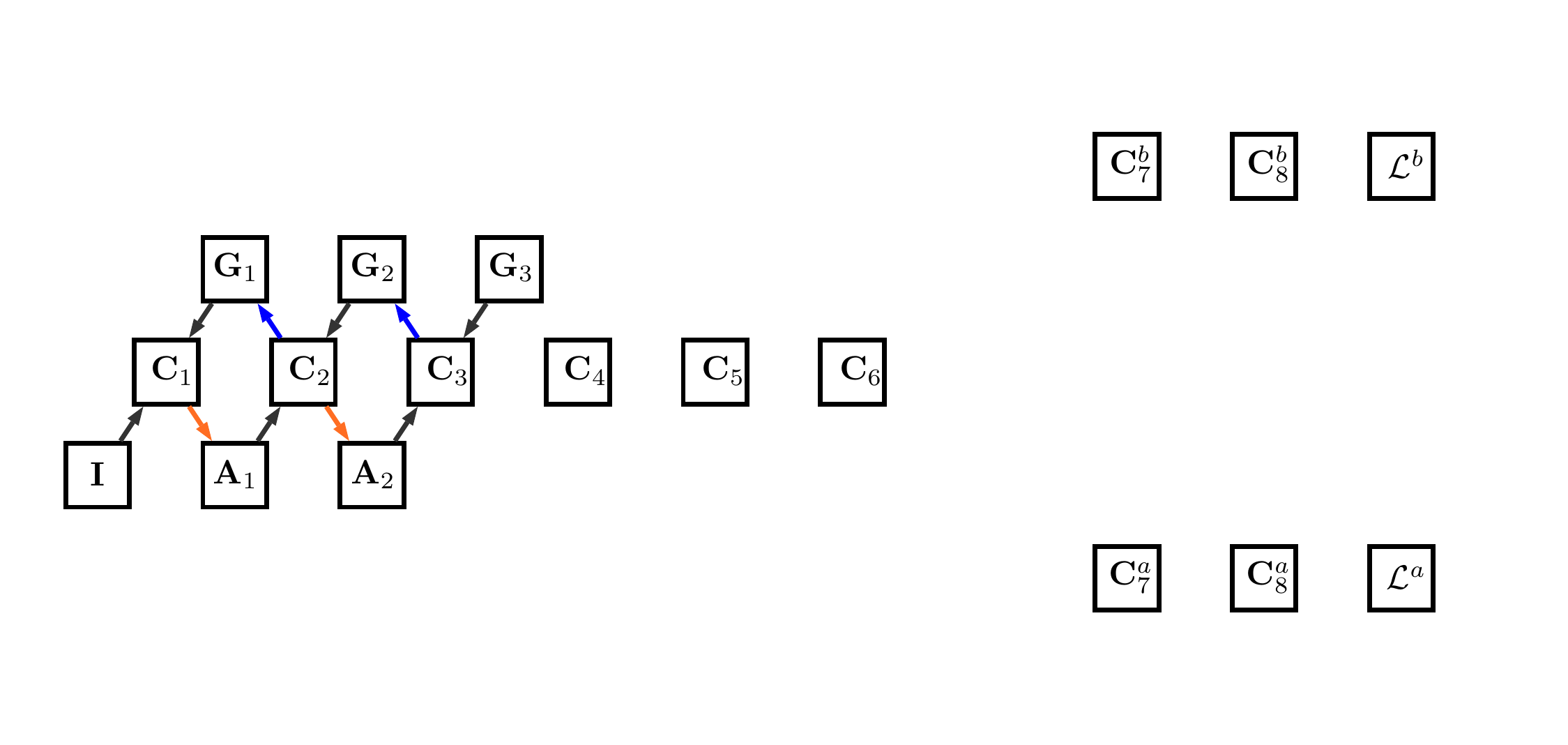}}
\end{figure}

\section{Memory-Bound Multi-Task Training}

\label{memory}

We now turn to handling memory limitations, which turns out to be a  major problem when  training a network for many tasks. In order to handle these problems we build on recent advances in memory-efficient backpropagation for deep networks \cite{GruslysMDLG16,ChenXZG16} and adapt them to the task of multi-task learning\footnote{I thank George Papandreou for suggesting this direction}. 
We start by describing the basic idea behind the algorithm of \cite{ChenXZG16}, paving the way for the presentation of our extension to multi-task learning.
 
The baseline implementation of the back-propagation algorithm maintains all intermediate layer activations computed  during the forward pass. As illustrated in \reffig{fig:vanillasingle}, 
during  the backward pass each layer then combines its stored activations  with the back-propagated gradients coming from the layer(s) above, finds the gradients for its own parameters, and then back-propagates  gradients to the layer(s) below.
While this strategy achieves computational efficiency by reusing the computed activation signals, it is memory-demanding, since it requires storing all intermediate activations. In the popular Caffe \cite{caffe} library memory is also allocated for all of the gradient signals, since a priori these could feed into multiple layers for a DAG network.

\newcommand{\nl}{L}


If we consider for simplicity that every layer requires $\nbytes$ bytes of memory for its activations, and gradient signals, and  we have a network with a total of $\nl$ layers,  the memory complexity of a naive implementation would be $2 \nbytes \nl$ - which can become prohibitive for large values of $\nl$.

The memory-efficient alternative described in \cite{ChenXZG16} is shown in \reffig{fig:memorysingle}. 
In a first step, shown in \reffig{fig:memorysingle}(a), we perform a 
first forward pass through the network where we store the activations of only a subset of the layers - for a network of depth $\nl$, $\sqrt{\nl}$ activations are stored, lying $\sqrt{\nl}$ layers apart, while the other intermediate activations, shown in grey, are discarded as soon as they are used. 
 Once this first stage is accomplished, we run $\sqrt{\nl}$ times backpropagation over sub-networks of length $\sqrt{\nl}$, as shown in \reffig{fig:memorysingle}(b)-(d). The stored activations help us start the backpropagation at a deeper layer of the network, acting like anchor points for the computation: any subnetwork requires the activation of its lowest level, and the gradient signal at its highest layer. 
 It can be seen that through this scheme the total complexity can be reduced from $\nl\nbytes$ to $2 \sqrt{\nl} \nbytes$, since we retain $\sqrt{\nl}$ activation signals, and at any step perform back-propagation over a subnetwork of length $\sqrt{\nl}$.

 This algorithm was originally introduced for chain-structured graphs; having described it, the adaptation to our case is straightforward. 

Considering that we have  $\nlcnn$ layers for the shared CNN trunk, $T$ tasks,  and $\nltask$ layers per task, the memory complexity of the naive implementation would be $2 \nbytes (\nlcnn   +  T \nltask )$, as can also be seen from \reffig{fig:vanillamtask}.

A naive application of the algorithm presented above would result in a reduction of memory complexity down to $2 \nbytes \sqrt{\nlcnn   +  T \nltask }$.
However, we realize that after the branching point of the different tasks (layer 6 for our figure), the computations are practially decoupled: each task-specific branch works effectively on its own, and then returns a gradient signal to layer 6. These gradient signals are accumulated over tasks, since our cost is additive over the task-specific losses. 

Based on this observation, we realize that the memory complexity can be reduced to be {\emph{independent}} of $T$: since each task can `clean up' all of the memory allocated to it, this results in a memory footprint of $2 \nbytes \sqrt{\nlcnn   +  \nltask}$ rather than  $2 \nbytes \sqrt{\nlcnn   +  T\nltask}$.

This has allowed us to load an increasing number of tasks on our network without encountering memory issues. Using a 12GB Nvidia card we have been able to use a three-layer pyramid, with the largest image size being 921x621, and using skip-layer connections for all network layers, pyramid levels, and tasks, for seven tasks. The largest dimension that would be possible without the memory-efficient option for our present number of tasks  would be 321x321 - and that would only decrease as more tasks are used.

Apart from reducing memory demands, we notice that we can also reduce computation time by performing a {\emph{lazy evaluation}} of the gradient signals accumulated at the branching point. In particular, if a training sample does not contain ground-truth for certain tasks, these will not contribute any gradient term to the common CNN trunk; as such the computation over task-specific branches can be avoided for an instance that does not contain ground-truth for the task. This results in a substantial acceleration of training (2- to 4-fold in our case), and would be essential to scale up training for even more tasks.

\mycomment{
Another major challenge has been the effect of image resolution on performance; we have observed that reducing the image resolution can have an adversarial effect on detection and semantic segmentation performance. However, training with high-resolution images can quickly result in GPU memory issues. We have engineered a two-stage method to train first a network that learns a common low-level, convolutional representation for all tasks, and then freezes the convolutional layers, so as to train in a decoupled manner the subsequent tasks at a higher resolution. 
}

\section{Experiments}


Our experimental evaluation has two objectives:
The first one is to show that the generic \ubernet architecture  introduced in \refsec{arch} successfully addresses a broad range of tasks.
In order to examine this we compare primarily to results obtained by methods that rely on the VGG network \cite{simonyan2014very} - more recent works e.g. on detection \cite{DaiLHS16} and semantic segmentation \cite{ChenPK0Y16} have shown improvements through the use of deeper ResNets \cite{HeZRS15}, but we consider the choice of network to be in a sense orthogonal to the goal of this section. 

The second objective is to explore how incorporating more tasks affects the  performance in the individual tasks. In order to remove erroneous sources of variation we use a common initialization for all single- and multi- task networks, obtained by pretraining a network for joint semantic segmentation and object detection, as detailed in \refsec{settings}. Furthermore, the multi-task network is trained with a union of datasets corresponding to the multiple tasks that we aim at solving. There we have used a particular proportion  of images per dataset, so as to moderately favor high-level tasks, as detailed in \refsec{settings}. Even though using a larger task-specific dataset may boost performance for the particular task, the single task networks are only trained with the subset of the multi-task dataset that pertains to the particular task. 
This may be sacrificing some performance with respect to competing methods, but ensures that the loss term pertaining to a task is unaffected by the single- versus multi-task training, and facilitates comparison. 



 \subsection{Experimental settings}
 \label{settings}
 \textbf{Optimization: } For all of the single-task experiments  we use SGD with a momentum of 0.9 and a minibatch size of 10 - with the exception of detection, where we use a minibatch size of 2, following \cite{Girshick15}.  For the multi-task experiments we use our asynchronous SGD algorithm with effective minibatch sizes of 2 for detection-related parameters, 10 for other task-specific parameters and 30 for the shared CNN features, as justified in \refsec{diverse}. 
 With the exception of the initialization experiment described right below, we always use 5000 iterations, starting with a learning rate of 0.001 and decrease the learning rate by a factor of 10 after 3000 iterations. Other optimization schemes will be explored in a future version of this work.

\textbf{Initialization of labelling and detection network:}
We  use a common initialization for all experiments, which requires having at our disposal parameters for both the convolutional labelling tasks, and the region-based detection task. We could use the ImageNet-pretrained VGG network for this, but exploiting pretraining on MS-COCO has been shown to yield boosts in performance e.g. in \cite{ChenPK0Y16,Girshick15}. Leaving a joint pretraining on MS-COCO for a future version of this work, we take a shortcut and instead form a `frankenstein' network  where we stitch together two distinct  variants of the VGG network, which have both been pretrained on MS-COCO. In particular we use the network of \cite{ChenPK0Y16} for semantic segmentation (`COCO-S') and the network of \cite{Girshick15} for detection, (`COCO-D'). 
 
The two-task network has (i) a common convolutional trunk, up to the fifth convolutional layer, (ii) a detection branch, combining an RPN and an SPP-Pooling layer followed by two fully-connected layers ($\mathrm{fc6,fc7}$), and (iii) a fully-convolutional branch, used for semantic segmentation. The fully-connected  branches  in (ii) and (iii) are initialized with the parameters of the respective pretrained networks, COCO-D, COCO-S, while for (i) we initialize the parameters of the common layers with the COCO-D parameters.  We finetune this network for 10000 iterations on the VOC07++ set \cite{Girshick15}  stands for  the union of PASCAL VOC 2007 trainval and PASCAL VOC 2012 trainval sets; we start with a learning rate of $0.001$ and decrease it to $0.0001$ after 6000 iterations.

\textbf{Datasets: } 
A summary of the datasets used in our experiments is provided in \reftable{numbers}. The 5100 images in the BSD dataset correspond to dataset augmentation of the 300 trainval images of BSD with 16 additional rotations. All of these numbers are effectively doubled with flipping-based dataset augmentation, while the VOC-related datasets are used twice, which amounts to placing a higher emphasis on the high-level tasks. 

We note that the VOC'12 validation is used for the evaluation of the human part segmentation, semantic boundary detection and saliency estimation tasks.  This  means that in general we report numbers for two distinct networks: one where the VOC2012 validation set is included during training, based on which we report results on detection and semantic segmentation; and one where VOC2012 validation is excluded from training, which gives us results on human parts, semantic boundaries, and saliency.  



\renewcommand{\yes}{All}
\renewcommand{\no}{0}
\newcommand{\hspt}{}
\begin{table}
	\begin{center}
			\resizebox{\linewidth}{!}{
				\begin{tabular}{l cccccc}
					&  VOC'07 	&  VOC'12  &  
					 VOC'12   &
					   NYU  & MSRA10K  & BSD  \\
					& trainval & train & val &  & & \\
					& 5011  & 5717 & 5823 & 23024 & 10000 & 5100 \\
					Detection & 5011 & 5717 & 5823 & \no & \no & \no \\
					S. Segmentation & 422 & 4998 & 5105 & \no & \no & \no\\
					S. Boundaries & \no & 4998 & 5105 & \no &  \no & \no \\
					Human Parts & \no & 1716  & 1817 & \no & \no & \no\\
					Normals &  \no & \no & \no & 23024 & \no & \no\\ 
					Saliency 	& \no & \no 	& \no &	\no & 10000 & \no\\
					Boundaries 	& \no & 4998 	& 5105 & \no & \no & 5100
				\end{tabular}
		}
		\caption{Datasets and numbers of images containing ground truth for the different tasks considered in this work.
			\label{numbers}}
	\end{center}
\end{table}

%

\newcommand{\resfig}[1]
{
	{\setlength{\tabcolsep}{.0em}	
		\begin{tabular}{cccccccc}
			Input & Boundaries & Normals & Depth & Saliency & S. Boundaries & S. Segmentation & Human Parts\\
			\includegraphics[width=\wdtg]{\rt/#1im}&
			\includegraphics[width=\wdtg]{\rt/#1bnd}&
			\includegraphics[width=\wdtg]{\rt/#1nrm}&
			\includegraphics[width=\wdtg]{\rt/#1dep}& 
			\includegraphics[width=\wdtg]{\rt/#1sal}&
			\includegraphics[width=\wdtg]{\rt/#1sbnd}&
			\includegraphics[width=\wdtg]{\rt/#1seg}&
			\includegraphics[width=\wdtg]{\rt/#1prt}
		\end{tabular}
	}
}

\newcommand{\wdh}{.124\linewidth}

\newcommand{\resline}[1]
{
	\includegraphics[width=\wdh]{\rt/l#1_input}&
	\includegraphics[width=\wdh]{\rt/l#1_boundaries}&	
	\includegraphics[width=\wdh]{\rt/l#1_surface_normals}&
	\includegraphics[width=\wdh]{\rt/l#1_saliency}&
	\includegraphics[width=\wdh]{\rt/l#1_semantic_boundaries}& 
	\includegraphics[width=\wdh]{\rt/l#1_semantic_segmentation}& 
	\includegraphics[width=\wdh]{\rt/l#1_human_parts}&
	\includegraphics[width=\wdh]{\rt/l#1_detection}\\
}

\newcommand{\resrow}[7]
{
	\includegraphics[width=\wdh,height=\hgh]{\rt/l#2_#1}&
	\includegraphics[width=\wdh,height=\hgh]{\rt/l#3_#1}&	
	\includegraphics[width=\wdh,height=\hgh]{\rt/l#4_#1}&
	\includegraphics[width=\wdh,height=\hgh]{\rt/l#5_#1}&
	\includegraphics[width=\wdh,height=\hgh]{\rt/l#6_#1}\\
} 

\newcommand{\hspl}{.1cm}

\newcommand{\tf}[1]{#1}

\newcommand{\ima}{26}
\newcommand{\imb}{7}
\newcommand{\imc}{8}
\newcommand{\imd}{4}
\newcommand{\ime}{16}
\newcommand{\imf}{19}
\renewcommand{\wdh}{.17\linewidth}
\newcommand{\hgh}{.13\linewidth}
\newcommand{\figcomment}[1]{#1}
\newcommand{\cmt}[2]{\rotatebox{90}{\hspace{#2}{\small #1}}}
\figcomment{
\begin{figure*}
	{	
		\begin{tabular}{cccccc}
				
		\cmt{Input}{20pt} & 	\resrow{input}{\ima}{\imb}{\imc}{\imd}{\ime}{\imf}
		\cmt{Boundaries}{10pt} & 	\resrow{boundaries}{\ima}{\imb}{\imc}{\imd}{\ime}{\imf}
		\cmt{Surface Normals}{2pt} &	\resrow{surface_normals}{\ima}{\imb}{\imc}{\imd}{\ime}{\imf}
		\cmt{Saliency}{20pt} & 	\resrow{saliency}{\ima}{\imb}{\imc}{\imd}{\ime}{\imf}
		&	\multicolumn{5}{c}{\includegraphics[width=.75\linewidth]{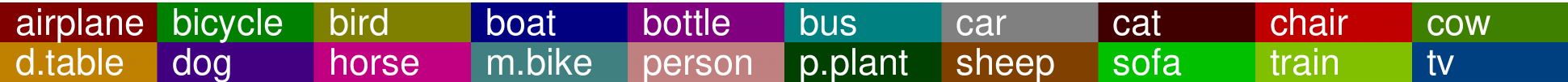}}\\
		\cmt{Sem.c Boundaries}{0pt} & 	\resrow{semantic_boundaries}{\ima}{\imb}{\imc}{\imd}{\ime}{\imf}
		\cmt{Sem.c Segmentation}{0pt} & 	\resrow{semantic_segmentation}{\ima}{\imb}{\imc}{\imd}{\ime}{\imf}
		\cmt{Object Detection}{0pt} &	\resrow{detection}{\ima}{\imb}{\imc}{\imd}{\ime}{\imf}		
		&	\multicolumn{5}{c}{\includegraphics[width=.75\linewidth]{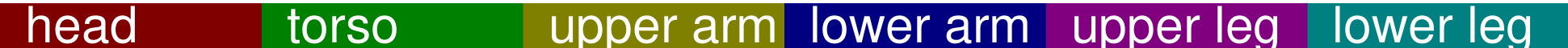}}\\
		\cmt{Human Parts}{10pt} & 	\resrow{human_parts}{\ima}{\imb}{\imc}{\imd}{\ime}{\imf}
		\end{tabular}}
		\caption{Qualtitative results of our network. Please note the human pictures  detected in the first two columns, as well as the range of scales  successfully handled by our network.}
	\end{figure*}
}

\renewcommand{\ima}{25}
\renewcommand{\imb}{12}
\renewcommand{\imc}{29}
\renewcommand{\imd}{33}
\renewcommand{\ime}{35}
\renewcommand{\imf}{33}
\renewcommand{\wdh}{.17\linewidth}
\renewcommand{\hgh}{.13\linewidth}
\figcomment{
	\begin{figure*}
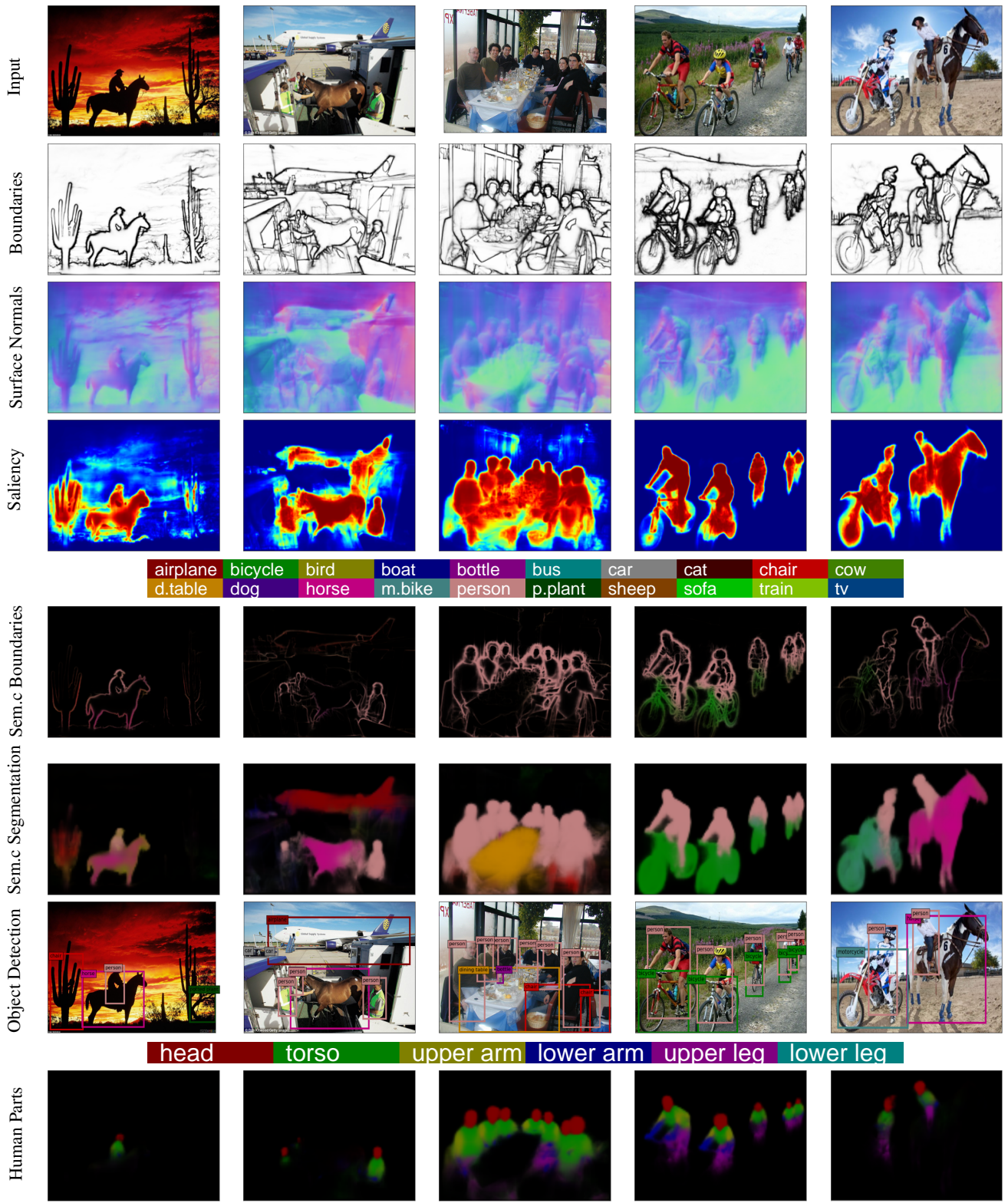

		{	
			
			\begin{tabular}{cccccc}				
				\cmt{Input}{20pt} & 	\resrow{input}{\ima}{\imb}{\imc}{\imd}{\ime}{\imf}
				\cmt{Boundaries}{10pt} & 	\resrow{boundaries}{\ima}{\imb}{\imc}{\imd}{\ime}{\imf}
				\cmt{Surface Normals}{2pt} &	\resrow{surface_normals}{\ima}{\imb}{\imc}{\imd}{\ime}{\imf}
				\cmt{Saliency}{20pt} & 	\resrow{saliency}{\ima}{\imb}{\imc}{\imd}{\ime}{\imf}
				&	\multicolumn{5}{c}{\includegraphics[width=.75\linewidth]{classbar}}\\
				\cmt{Sem.c Boundaries}{0pt} & 	\resrow{semantic_boundaries}{\ima}{\imb}{\imc}{\imd}{\ime}{\imf}
				\cmt{Sem.c Segmentation}{0pt} & 	\resrow{semantic_segmentation}{\ima}{\imb}{\imc}{\imd}{\ime}{\imf}
				\cmt{Object Detection}{0pt} &	\resrow{detection}{\ima}{\imb}{\imc}{\imd}{\ime}{\imf}		
				&	\multicolumn{5}{c}{\includegraphics[width=.75\linewidth]{humanbar}}\\
				\cmt{Human Parts}{10pt} & 	\resrow{human_parts}{\ima}{\imb}{\imc}{\imd}{\ime}{\imf}
			\end{tabular}}
			\caption{Qualtitative results, continued: please note that the leftmost image has practically no color information, which can justify the mistake of the semantic segmentation and object detection tasks: the left cactus  is incorrectly labelled as a chair, apparently mistaken for a thorny throne.}
		\end{figure*}
	}

\subsection{Experimental Evaluation}
\textbf{Object Detection:}
%
We start by verifying in the `Ours 1-Task' row that we can replicate the results of \cite{Girshick15}; exceptionally for this experiment, rather than using the initialization described above,   we start from the MS-COCO
pretrained network of \cite{Girshick15}, finetune on the VOC 2007++ dataset,  and test on the VOC 2007 test dataset. The only differences are that 
we use a minimal image side of 621 rather than 600, a maximal side of 961 rather than 1000, so as to comply with the $32k+1$ restriction on dimensions of \cite{ChenPK0Y16}, and use  convolution with holes followed by appropriately modified RPN and ROI-pooling layers to get effectively identical results as \cite{Girshick15}. Adding holes to the RPN network did not seem to help (not reported). 

In the following row we measure the performance of the network obtained by training for the joint segmentation and detection task, which as mentioned in \refsec{settings} will serve as our starting point for all ensuing experiments. 
\mycomment{ - where we use again the MS-COCO pretrained network for all of the detection-related layers, but append  `convolutionalized' $\mathrm{fc_6,fc_7,fc_8}$ layers on top of the VGG network so as to obtain semantic segmetnation masks. As mentioned above these are initialized with the `COCO-S' network parameters.}
  After finetuning on VOC2007++ we observe that we actually get a small boost in performance, which is quite promising, since it is likely to be telling us that the additional supervision signal for semantic segmentation helped the detection sub-network learn something better about detection. 
 
However when increasing the number of tasks  performance drops, but is still comparable to the strong baseline of \cite{Girshick15}. As we will see in the \refsec{mtask} this is not necessarily obvious - a difference choice of task weight parameters  can adversely influence detection performance while favoring  other tasks. 

\begin{table}[!h]

		\begin{tabular}{|l|c|}
			\hline
		Method	&	mAP \\\hline\hline
			F-RCNN, \cite{Girshick15} VOC 2007++  & 73.2\\\hline
			F-RCNN, \cite{Girshick15} MS-COCO +  VOC 2007++  & 
			78.8\\\hline	
			Ours, 1-Task & 78.7 \\\hline
			Ours, 2-Task & 80.1 \\\hline
			Ours, 7-Task & 77.8 \\\hline
		\end{tabular}
	\caption{Mean Average Precision performance (\%) on the PASCAL VOC 2007 test set.}
	\label{table:results_ap}
\end{table}

\textbf{Semantic Segmentation:} The second task that we have tried is semantic segmentation. Even though a really broad range of techniques have been devised for the problem (see e.g. review in \cite{ChenPK0Y16} for a recent comparison), we only compare to the  methods lying closest to our own, which in turns relies on the `Deeplab-Large Field of View (FOV)' architecture of \cite{Chen2015iclr}. We remind that, as detailed in \refsec{arch}, we deviate from the Deeplab architecture by using linear operations on top of skip layers, by using a multi-scale architecture, and by not using any DenseCRF post-processing.
 
\begin{table}[!h]
	\begin{tabular}{|l|c|c|c|}
		\hline
	Method	& mean IoU \\\hline\hline
		Deeplab -COCO + CRF \cite{papa15}& 70.4\\\hline
		Deeplab Multi-Scale \cite{iclr16} & 72.1 \\
		Deeplab Multi-Scale -CRF \cite{iclr16} & 74.8 \\\hline
		Ours, 1-Task  & 72.4\\
		Ours, 2-Task  & 72.3\\
		Ours, 7-Task & 68.7 \\\hline
	\end{tabular}
	\caption{Semantic segmentation  - mean Intersection Over Union (IOU) accuracy on PASCAL VOC 2012 test.\label{table:results_e}}
\end{table}
We first observe that thanks to the use of multi-scale processing we get  a similar improvement over the single-scale architecture as the one we had obtained in \cite{iclr16}.
Understandably this ranks below the latest state-of-the-art results, such as the ones obtained e.g. in \cite{ChenPK0Y16} with ResNets and Atrous Spatial Pyramid Pooling; but these advances are complementary and easy to include in our network's architecture. 

Turning to the results of the two-task architecture, we observe that quite surprisingly we get effectively the same performance. This is not obvious at all, given that for this two-task network our starting point has been a VGG-type network that uses the detection network parameters up to the fifth convolutional layer, rather than the segmentation parameters. Apparently, after 10000 iterations of fine-tuning the shared representation was modified to be appropriate for the semantic segmentation task.

Turning to the multi-task network performance, we observe that performance drops as the number of tasks increases. Still, even without using CRF post-processing, we fare comparably to a strong baseline, such as \cite{papa15}.


 \textbf{Human Part Segmentation:}  This task can be understood as a special case of  semantic segemntation, where now we aim at assigning human part labels. Recent work has shown that semantic part segmentation is one more task that can be solved by CNNs \cite{TsogkasSemanticPart15,HAZN,ChenYWXY15,LiangSFLY16,ChenPK0Y16}.

 \begin{table}[!h]
 	\begin{tabular}{|l|c|c|c|}
 		\hline
 		Method & mean IoU \\\hline\hline
 		Deeplab LargeFOV \cite{HAZN} & 51.78\\\hline
 		Deeplab LargeFOV-CRF\cite{HAZN} & 52.95 \\\hline
 		Multi-scale averaging \cite{ChenYWXY15} & 54.91 \\\hline
 		Attention \cite{ChenYWXY15} & 55.17\\\hline
 		Auto Zoom  \cite{HAZN} & 57.54\\\hline
 		Graph-LSTM \cite{LiangSFLY16} & 60.16\\\hline
 		Ours, 1-Task & 51.98\\\hline
 		Ours, 7-Task & 48.82 \\\hline
 	\end{tabular}
 	\caption{Part segmentation - mean Intersction-Over-Union accuracy on the dataset of \cite{chen_cvpr14}. \label{table:results_e}}
 \end{table}
 
 We use the dataset introduced in \cite{chen_cvpr14} and train a network that is architecturally identical to the one used for semantic segmentation, but is now finetuned for the task of segmenting human parts. As a general comment on this task we can observe that here structured prediction yields  quite substantial improvements, apparently due to the highly confined structure of the output space in this labelling task. Since we do not use any such post-processing yet, it is only fair to compare to the first, CRF-free method, and leave an integration with structured prediction for future work.
 
 As can be seen in \reftable{table:results_e} for the single-task case we perform comparably. 
 However,  for the multi-task case performance drops substantially (more than 10$\%$). This  may be due to the scarcity of data that contain annotation for the task: when training the single task network all of the data contain annotations for human parts, while when training the multi-task network we have human part annotations in only 3432 out of  the 59552 images used to train the whole network (referring to \reftable{numbers}, we remind the reader that we use twice the VOC-related datasets).
One potential remedy is to increase the weight of the task's loss, or the learning rates of the task-specific parameters, so that the parameter updates are more effective; another alternative is to give the multi-task network more training iterations, so that we pass more times over the part annotations. We are exploring these options. 
 

\textbf{Semantic Boundary Detection:}
 We evaluate our method on the Semantic Boundary Detection task defined in 
 \cite{hariharan2011semantic}, where the goal is to find where instances of the 20 PASCAL classes have discontinuities. This can be understood as a combination of semantic segmetnation and boundary detection, but can be tackled head-on with fully convolutional networks. 
 We train on the VOC2012 train and evaluate on VOC2012 val.
 
 \begin{table}[!h]
 	\begin{tabular}{|l|c|c|} \hline
 		Method			&  mAP & mMF 	 		
 		\\\hline	\hline
 		Semantic Contours \cite{hariharan2011semantic}  & 20.7		&   28.0  	\\\hline
 		Situational Boundary \cite{UijlingsF15} & 31.6 & -  \\\hline
 		High-for-Low \cite{BertasiusST15} 	&  47.8  &  58.7 \\\hline
 		High-for-Low-CRF \cite{BertasiusST15} & 54.6 & 62.5 \\\hline
 		Ours, 1-Task & 54.3 & 59.7 \\\hline
 		Ours, 7-Task & 44.3 & 48.2 \\\hline
 	\end{tabular}
 	\caption{Semantic Boundary Detection Results: we report mean Average Precision (AP) performance (\%) and  Mean Max F-Measure Score on the validation set of PASCAL VOC 2010, provided by \cite{hariharan2011semantic}.}
 	\label{table:results_ap}
 \end{table}
 
 We compare to the original method of \cite{hariharan2011semantic}, the situational boundary detector of \cite{UijlingsF15}, and the High-for-Low method of \cite{BertasiusST15}. The authors of \cite{BertasiusST15} go beyond the individual task of boundary detection and explore what gains can be obtained by providing as inputs to this task the results of a separate semantic segmentation system (`High-for-Low-CRF' row). Even though combining the outputs of different tasks is one of our immediate next goals, we do not consider it yet here. Still, we observe that even applying our architecture out-of-the-box we get reasonably close results, and substantially better than their standalone semantic boundary detection result. Performance deteriorates for the multi-task case, but remains quite close to the current `standalone' state-of-the-art.

 \textbf{Boundary Detection:} We  train our network on the union of the (dataset-augmented) BSD trainval set and boundary images from the VOC context dataset \cite{context} and evaluate it on the test set of the Berkeley Segmentation Dataset (BSD)  of \cite{MFTM01} and. We compare our method to some of the best-established methods of boundary detection \cite{berkeley11,sfs}, as well as more recent, deep learning-based ones \cite{aistat,ganin2014n,shi15,iclrbnd,shen2015deepcontour,hed,iclr16}. 
 
 \begin{table}[!h]
 	\begin{tabular}{|@{}l@{}|c|c|c|}
 		\hline
 		Method  & ODS & OIS  & AP \\
 		\hline
 		\hline
 		gPb-owt-ucm \cite{berkeley11}   &  0.726 & 0.757 & 0.696 \\\hline
 		SE-Var        \cite{sfs} &  0.746 & 0.767 & 0.803 \\\hline
 		DeepNets      \cite{aistat} &  0.738 & 0.759 & 0.758 \\\hline
 		N4-Fields     \cite{ganin2014n} &  0.753 & 0.769 & 0.784 \\\hline
 		DeepEdge      \cite{shi15} &  0.753 & 0.772 & 0.807 \\\hline
 		CSCNN         \cite{iclrbnd} &  0.756 & 0.775 & 0.798 \\\hline
 		DeepContour  \cite{shen2015deepcontour}  &  0.756 & 0.773 & 0.797 \\\hline
 		HED-fusion  \cite{hed} &  0.790 & 0.808 & 0.811 \\
 		HED-late merging  \cite{hed} &  0.788 & 0.808 & 0.840 \\
 		\hline
 		Multi-Scale\cite{iclr16}  & 0.809 & 0.827 & 0.861\\
 		Multi-Scale  +sPb	\cite{iclr16} & 0.813 & 0.831 & 0.866\\\hline
 		Ours, training setup of \cite{{iclr16}}  & 0.815   &  0.835 & 0.862 \\\hline
 		Ours, 1-Task  & 0.791  &  0.809 & 0.849 \\
 		Ours, 7-Task & 0.785  & 0.805 &  0.837 \\
 		\hline
 	\end{tabular}
 	\caption{Boundary Detection results: we report the maximal F meaure obtained at the Optimal Dataset Scale, Optimal Image Scale, as well as the Average Precision on the test set of the BSD dataset \cite{MFTM01}.  \label{table:hedfinal}}
 \end{table}
 
 A  first experiment has been to train our new network with the exact same experimental setup as the one we had used in \cite{iclr16} - including Graduated Deep Supervised Network training,  a mix of 30600 images obtained by dataset augmentation from the BSD (300 images augmented by 3 scales, 2 horizontal flips, and 16 rotations) with 20206 images from VOC-context (10103 images with two horizontal flips). The differences  are that we now use batch normalization, which allows us to increase the layer-specific learning rates up to 10, and also that we now use the  `convolutionalized' fully-connected layers of the VGG network. The improvement in performance is quite substantial: the maximal F-measure increases from 0.809 to 0.815, surpassing even the performance we would get in \cite{iclr16} by using spectral boundaries on top. 
 
 Still, these settings (graduated DSN, high learning rates) were not as successful on the remaining tasks, while using a mix of data where the images of BSD are three times more than the images of VOC would skew the performance substantially in favor of the low-level task of boundary detection - since the training objective is clearly affected by the number of images containing ground truth for one task versus the other.
 
 We therefore remove the side losses for the skip layers,  reduce the layer-specific learning rate to 1 (which is still quite higher than the 0.001 layer-specific rate used in \cite{hed,iclr16}), and use the particular mix of data used to train the \ubernet in the multi-task setup. This means that, after left-right flipping, we use 10200 boundary samples from BSD (i.e. no scale augmentation) and 20206 samples from VOC. 
 
 As shown in the `Ours, 1 Task' row of 
 \reftable{table:hedfinal} this can substantially affect performance  - but we still  remain competitive to previous state-of-the-art works, such as \cite{hed}. For the multi-task training case performance drops a bit more, but always stays at a reasonably good level when compared to standard strong baselines such as \cite{hed}.  
 
 
 \textbf{Saliency Estimation:} We train  on the MSRA-10K dataset of \cite{WangLLLS15} and evaluate on the PASCAL-S dataset of \cite{LiHKRY14}, where a subset of the Pascal VOC10 validation set is annotated. Additional datasets are typically used to benchmark this task, e.g. in \cite{saliencycvpr16}, we will explore the performance of our method on those datasets in the future.
 
 We only use flipping for dataset augmentation during training. We compare to some classic methods \cite{PerazziKPH12,msra,JiangLYP13} as well as more recent ones \cite{QinLXW15,WangLRY15,ZhaoOLW15,saliencycvpr16} that typically rely on deep learning. We note that our method sets a new state-of-the-art for this dataset, and even for the multi-task training case, our method 
 outperforms the previous state-of-the-art which was the CRF-based variant of \cite{saliencycvpr16}.  
 
 \begin{table}[!h]
 		\begin{tabular}{|l |c|}		
 			\hline	
 		Method	&  MF \\ \hline\hline
 	SF \cite{PerazziKPH12} & 	0.493\\\hline
GC \cite{msra} & 0.539\\\hline
DRFI \cite{JiangLYP13} & 0.690\\\hline
PISA \cite{WangLLLS15}& 0.660\\\hline
BSCA \cite{QinLXW15} & 0.666\\\hline
			LEGS \cite{WangLRY15} & 0.752 \\\hline
			MC  \cite{ZhaoOLW15} & 0.740 \\\hline
			MDF  \cite{LiY15} & 0.764\\\hline
			FCN  \cite{saliencycvpr16} & 0.793 \\
 			DCL   \cite{saliencycvpr16} & 0.815 \\
 			DCL + CRF \cite{saliencycvpr16} & 0.822 \\\hline
 			Ours, 1-Task & 0.835 \\
 			Ours, 7-Task & 0.823 \\\hline
 		\end{tabular}
 	\caption{Saliency estimation results: we report the Maximal F-measure (MF)  on the PASCAL Saliency dataset of \cite{LiHKRY14}.\label{table:results_e}}
 \end{table}

 \textbf{Surface Normal Estimation:} For this task surface normals are typically estimated from point cloud data, rather than directly measured. Both the training, and also the evaluation of a normal estimation algorithm may therefore be affected by this step.  
 We train on the training set if \cite{nyu} where normals are estimated by   \cite{LadickyZP14} and extend it with 20K images of normal ground truth estimated from the raw images in the training scenes of \cite{nyu}; since the method of \cite{LadickyZP14} is not publicly available, we use as a surrogate the method of \cite{Ren12}. Competing methods, e.g. \cite{Eigen15,BansalRG16} are using alternative normal estimation methods for the extended data, but we would not expect the differences because of this to be too large. 
 
 \newcommand{\hsp}{\hspace{5pt}}
 \newcommand{\hsps}{\hspace{2pt}}
  \begin{table}[!h]
 	\begin{center}

 			\begin{tabular}{|l|c @{\hsps} c @{\hsps} c @{\hsp} c @{\hsp} c|}
 				\hline		 
 			Method	&  Mean &  Median &  $11.25^{\circ}$ & $22.5^{\circ}$ & $30^{\circ}$  \\\hline\hline
				 VGG-Cascade \cite{Eigen15} &  22.2 & 15.3& 38.6 & 64.0 & 73.9\\\hline
 				VGG-MLP \cite{BansalRG16} & 	19.8 & 12.0  & 47.9 & 70.0 & 77.8 \\\hline
 				VGG-Design \cite{WangFG15} & 26.9 & 14.8 & 42.0 & 61.2 & 68.2 \\\hline
 				VGG-fused \cite{WangFG15} & 27.9 & 16.6 & 37.4  & 59.2 & 67.1 \\\hline
 				Ours, 1-Task $\gamma=50$	&	21.4 & 15.6 &  35.3 & 65.9 & 76.9\\\hline
 				Ours, 1-Task $\gamma=10$  & 23.2 & 17.0 & 32.5 & 62.0 & 73.5\\\hline
 				Ours, 1-Task $\gamma=5$	& 23.3 & 17.6 & 31.1 & 60.8 & 72.7\\\hline
 				Ours, 1-Task $\gamma=1$ &  23.9 		& 18.1  & 29.8 & 59.7 & 71.9\\\hline
 				Ours, 7-Task 	& 26.7 & 22.0  & 24.2 & 52.0 & 65.9\\\hline
 			\end{tabular}

 	\end{center}
 	\caption{Normal Estimation on  NYU-v2  using the ground truth of \cite{LadickyZP14}. We report the Mean and Median Angle distance (in radians) and the percentage of valid pixels being with 11.25, 22.5, and 30 degrees of the ground-truth.}
 \end{table}
 
 We report multiple results for the single-task training case, obtained by setting values to the weight $\gamma$ of the loss term in \refeq{eq:objective}. We observe that this can have a quite substantial impact on performance. When setting a large weight we can directly compete with the current state-of-the-art, while a low weight can reduce performance substantially. As we will see however in the following subsection, it becomes necessary to set a reasonably low weight, or else this may have adverse effects on the performance of the remaining tasks. When using that lower weight, we witness a further drop in performance for the multi-task case.

Even though our multi-task network's performance  is not too different from the plain CNN-based result of \cite{WangFG15}, it is clear that there we have a somewhat unique gap in performance, when compared to what we seen in the remaining tasks. 

 Our conjencture is that this may be due to the geometric, and continuous nature of this task, which is quite different from the remaining labelling tasks. It may be that both the intermediate and final features of the VGG network are not  appropriate for this task `out-of-the-box', and it takes substantially large-scale modifications to the inner workings of the network (corresponding to a large weight on the task-specific loss) until the nonlinearities within the VGG network can accommodate the task. It is however interesting that both competing methods (VGG-MLP \cite{BansalRG16}, VGG-cascade \cite{Eigen15}) address the task by using additional layers on top of the VGG network (a Multi-Layer Perceptron in \cite{BansalRG16}, a coarse-to-fine cascade in \cite{Eigen15}). Even though in this work we have limited ourselves to using linear functions on top of the skip layers for the sake of simplicity and efficiency, these successes suggest that adding instead nonlinearities could be a way of improving performance for this task, as well as potentially also for other tasks.

\begin{table*}
	\resizebox{.95\textwidth}{!}{
		\begin{tabular}{|l|c|c c c|c|c|c|c|c|c|c|c|c|}
			\hline
			& Detection  		& \multicolumn{3}{c|}{	Boundaries }	& Saliency &  Parts & \multicolumn{3}{c|}{Surface Normals}  	& 	\multicolumn{2}{c|}{S. Boundaries }	  & S. Segmentation  \\\hline
			& mAP  
			& ODS & OIS & AP&  MF & mIoU & $ 11.2^{\circ}$ & $ 22.5^{\circ}$ &$ 30.0^{\circ}$ & MF & mAP & 
			mIoU\\\hline
			$\gamma = 1$			& 77.8  & 0.785 & 0.805 & 0.837 & 0.822 & 48.8 &  24.2 & 52.0 & 65.9 &  44.3 & 48.2 & 68.7 \\\hline
			$\gamma=5$			    & 76.2 &  0.779 & 0.805 & 0.836 & 0.820 & 36.7 & 23.1 & 51.0 & 64.9&  33.6 & 34.2  & 67.2 \\\hline
			$\gamma = 50$ 			&  73.5 &  0.772 & 0.802 & 0.830 & 0.814 &  34.2 &  27.7 & 57.3 & 70.2 &  28.6 & 33.2 & 63.5\\\hline
		\end{tabular}
	}
	\caption{Impact of the weight used for the normal estimation loss, when training for 7-tasks: Improving normal estimation comes at the cost of  decreasing  performance in the remaining tasks  (higher is better for all tasks). \label{table:resultsnrm}}
\end{table*}

\subsection{Effect of task weights}
The performance of our network on the multitude of task it adresses depends on the weights assigned to the losses of different tasks in \refeq{eq:objective}. 
If the weight  of one task is substantially larger, one can expect that this will skew the internal representation of the network in favor of the particular task, while negecting others.

\mycomment{
\begin{table*}[!h]
	\resizebox{\textwidth}{!}{
		\begin{tabular}{|l|c|c|c|c|c|c|c|c|c|c|}
			\hline
			& Det  		& 	Edg 	& Sal &  Prt & Nrm1  & Nrm2 & Nrm3 	& 	Sbd 	  & SegVal  	\\\hline
			lon-2.5 lob-5 			& 77.22 &  0.779 / 0.805 / 0.836 & 0.8200 & 48.659 & 0.231590 & 0.510119 & 0.649413&  24.66/ 0.34  & 74.418 \\\hline
			lon-5 lob-2 			&  77.42 &  0.780/ 0.805 /0.837  & 0.8223 & 48.805 &  0.216529 & 0.487322 & 0.627799  &   15.69 / 0.27 & 75.546 \\\hline
			lon-5 lob-5 			& 77.68  & 0.780 / 0.805 / 0.837 & 0.8220 & 48.608 &  0.217131 & 0.487126 & 0.627416 &  25.32 /0.34 & 75.589\\\hline
			10x  lon-2.5 lob-10 	& 76.34 & 0.787/0.810/0.849 & 0.8183 & 50.510 &  0.229456 & 0.508059 & 0.645994  & 41.00 /0.47 & 69.439\\\hline
			trainval repd-2 nrd lon-10 	&  77.42 & 0.784/ 0.808/ 0.843 & 0.8317 & 51.781 &  0.201141 & 0.465241 & 0.606101 &  37.64/ 0.43  & 75.266\\\hline
			trainval repd-3 lon-10  & 77.20 &  0.784 / 0.807 / 0.843 & 0.8346 & 52.199 &  0.180932 & 0.442156 & 0.585924 &  39.76 F 0.45 & 74.561 \\\hline
			10 tv-lon 10 repd2 & 75.58 & 0.791 / 0.812 / 0.854 & 0.8213 & 53.676 &  0.177990 & 0.441721 & 0.587608  & 47.18 /0.52 & 69.687\\\hline
			10 6K 			&  73.53 &  0.789 / 0.812 / 0.851 & 0.8146 &  51.741 &  0.277381 & 0.573961 & 0.702040 &  28.62 / 0.36 & 69.174\\\hline
		\end{tabular}
	}
	\caption{Effect of increasing the weight for the loss corresponding to the semantic boundary detection task. \label{table:resultsbd}}
\end{table*}
}
\label{mtask}

Motivated by the empirical results in the previous paragraphs we have explored the impact of modifying the weight attributed to  the normal estimation task in \refeq{eq:objective} in the case of solving multiple, rather than individual tasks. 
In \reftable{table:resultsnrm} we report how performance changes when we increase the weight of the normal estimation task  (previous experiments  relied on the $\gamma=1$ option). 

We realize that, at least for our particular experimental settings, there is `no free lunch', and the performance measures of the different tasks  act like communicating vessels.
 The evaluation may arguably be affected by our optimization choices; using e.g. larger batch sizes, or more iterations and a polynomial schedule as in \cite{ChenPK0Y16} could help. But the present results  indicate  that the common CNN trunk has apparently a bounded learning capacity, and suggests that inserting more parameters, potentially through additional nonlinear layers on top of the skip layers, may be needed to maintain high performance across all tasks. We will explore these directions in the future, as well as whether this effect persists when working with deeper networks such as ResNets.


\mycomment{
	\begin{table*}
		\begin{tabular}{|c c|c|c|c|c|c|c|c|}
			\hline
			&Date 		& 	Nrm 	& 	Sbd 	& Edge 		& Part 		& Saliency  & SegVal 	& Detection  	\\\hline
			Best 			& 14 Aug 	& 	57.9	& 	43.6 	& 0.780		& 48.833  	& 0.8278	& 75.896  	& 0.758  		\\\hline
			Validation 		& 14 Aug 	& 	59.5 	& 	47.7 	& 0.777		& 47.679 	& 0.8262 	& 75.683 	& 0.766			\\\hline
			no Aug  		& 5 Aug  	& 	60.6	& 	37.6 	& 0.784 	& 51.781	& 0.8317	& 75.266 	& 0.774 		\\\hline
			repd x 3 		& 23 Jul 	& 	58.6	&	39.8 	& 0.784 	& 52.199 	& 0.8346	& 74.561 	& 0.772			\\\hline
			fct 10 	(gohan)	& 5 Aug		& 	58.8	&	47.2  	& 0.791 	& 53.676 	& 0.8213	& 69.687	& 0.756 		\\\hline
			nrm10 + fct10 	& 3 Aug		& 	70.2	& 	28.6 	& 0.789 	& 51.741 	& 0.8146 	& 69.174 	& 0.7353		\\\hline
			kpt 			& 14 Aug	& 	58.2	&	47.1	& 0.777		& 49.29		& 0.8299	& 75.169	& 0.759			\\\hline	
		\end{tabular}	
		\caption{Overall results \label{table:results_e}}
	\end{table*}
}


\mycomment{
	\begin{figure*}
			{\setlength{\tabcolsep}{.1em}	
			\begin{tabular}{cccccccc}
		
		\resline{4}
		\resline{7}
		\resline{8}
		\resline{12}
		\resline{16}
				\resline{19}
	\resline{21}
		\resline{25}
			\resline{26}	
							\resline{28}
								\resline{33}
								\resline{30}

		\end{tabular}
	}
	\end{figure*}
}

\section{Conclusions and Future Work}
In this work we have introduced two techniques that allow us to train a  CNN that tackles a broad set of computer vision problems in a unified architecture. We have shown that one can effectively scale up to  many and diverse tasks, since 
the memory complexity is independent of the number of tasks, and incoherently annotated datasets can be combined during training.


There are certain straightforward directions for future work: (i) considering more tasks, such as symmetry, human landmarks, texture segmentation, or any other of the tasks indicated in the introduction (ii) using deeper architectures, such as ResNets \cite{HeZRS15} (iii) combining the dense labelling results with structured prediction
\cite{Adelaide,ChenPK0Y16,crfrnn,ChandraK16}. Research in these directions is underway, but, more importantly, we consider this work to be a first step in the direction of jointly tackling multiple tasks by exploiting the synergy between them - this has been a recurring theme in computer vision, e.g.  for integrating segmentation and recognition \cite{Keeler90,mumf94b,BottouBL97,TCYZ03,KoMa05,KTZ05,MaireYP11}, and we believe that successfully addressing this it is imperative to have a single network that can succesfully handle all of the involved tasks. 
The code for this work will soon be made publicly available from \url{http://cvn.ecp.fr/ubernet/}.


\section{Acknowledgements}
This work has been supported by the FP7-RECONFIG, FP7-MOBOT, and H2020-ISUPPORT EU projects, and equipment donated by NVIDIA. 
I thank George Papandreou for pointing out how  low-memory backpropagation can be implemented,  Pierre-Andr\'e Savalle for showing me how to handle prototxt files, Ross Girshick for making the Faster-RCNN system  publicly available, and Nikos Paragios for creating the environment where this work took place.


\begin{thebibliography}{100}\itemsep=-1pt
	
	\bibitem{berkeley11}
	P.~Arbelaez, M.~Maire, C.~Fowlkes, and J.~Malik.
	\newblock Contour detection and hierarchical image segmentation.
	\newblock {\em {PAMI}}, 2011.
	
	\bibitem{BansalRG16}
	A.~Bansal, B.~Russell, and A.~Gupta.
	\newblock Marr revisited: 2d-3d alignment via surface normal prediction.
	\newblock In {\em Proc. {CVPR}}, 2016.
	
	\bibitem{BelagiannisZ16}
	V.~Belagiannis and A.~Zisserman.
	\newblock Recurrent human pose estimation.
	\newblock {\em CoRR}, abs/1605.02914, 2016.
	
	\bibitem{BellUSB15}
	S.~Bell, P.~Upchurch, N.~Snavely, and K.~Bala.
	\newblock Material recognition in the wild with the materials in context
	database.
	\newblock In {\em Proc. {CVPR}}, 2015.
	
	\bibitem{bell16ion}
	S.~Bell, C.~L. Zitnick, K.~Bala, and R.~Girshick.
	\newblock Inside-outside net: Detecting objects in context with skip pooling
	and recurrent neural networks.
	\newblock {\em Proc. {CVPR}}, 2016.
	
	\bibitem{shi15}
	G.~Bertasius, J.~Shi, and L.~Torresani.
	\newblock Deepedge: {A} multi-scale bifurcated deep network for top-down
	contour detection.
	\newblock In {\em Proc. {CVPR}}, 2015.
	
	\bibitem{BertasiusST15}
	G.~Bertasius, J.~Shi, and L.~Torresani.
	\newblock High-for-low and low-for-high: Efficient boundary detection from deep
	object features and its applications to high-level vision.
	\newblock In {\em Proc. {ICCV}}, 2015.
	
	\bibitem{BilenV16}
	H.~Bilen and A.~Vedaldi.
	\newblock Integrated perception with recurrent multi-task neural networks.
	\newblock In {\em Proc. {NIPS}}, 2016.
	
	\bibitem{BottouBL97}
	L.~Bottou, Y.~Bengio, and Y.~LeCun.
	\newblock Global training of document processing systems using graph
	transformer networks.
	\newblock In {\em Proc. {CVPR}}, 1997.
	
	\bibitem{BourdevMM11}
	L.~D. Bourdev, S.~Maji, and J.~Malik.
	\newblock Describing people: {A} poselet-based approach to attribute
	classification.
	\newblock In {\em Proc. {ICCV}}, 2011.
	
	\bibitem{ChandraK16}
	S.~Chandra and I.~Kokkinos.
	\newblock Fast, exact and multi-scale inference for semantic image segmentation
	with deep gaussian crfs.
	\newblock In {\em Proc. {ECCV}}, 2016.
	
	\bibitem{ChenBPMY15}
	L.~Chen, J.~T. Barron, G.~Papandreou, K.~Murphy, and A.~L. Yuille.
	\newblock Semantic image segmentation with task-specific edge detection using
	cnns and a discriminatively trained domain transform.
	\newblock In {\em Proc. {CVPR}}, 2016.
	
	\bibitem{Chen2015iclr}
	L.~Chen, G.~Papandreou, I.~Kokkinos, K.~Murphy, and A.~L. Yuille.
	\newblock Semantic image segmentation with deep convolutional nets and fully
	connected crfs.
	\newblock In {\em Proc. {ICLR}}, 2015.
	
	\bibitem{ChenPK0Y16}
	L.~Chen, G.~Papandreou, I.~Kokkinos, K.~Murphy, and A.~L. Yuille.
	\newblock Deeplab: Semantic image segmentation with deep convolutional nets,
	atrous convolution, and fully connected crfs.
	\newblock {\em CoRR}, abs/1606.00915, 2016.
	
	\bibitem{ChenYWXY15}
	L.~Chen, Y.~Yang, J.~Wang, W.~Xu, and A.~L. Yuille.
	\newblock Attention to scale: Scale-aware semantic image segmentation.
	\newblock In {\em Proc. {CVPR}}, 2015.
	
	\bibitem{ChenXZG16}
	T.~Chen, B.~Xu, C.~Zhang, and C.~Guestrin.
	\newblock Training deep nets with sublinear memory cost.
	\newblock {\em CoRR}, abs/1604.06174, 2016.
	
	\bibitem{chen_cvpr14}
	X.~Chen, R.~Mottaghi, X.~Liu, S.~Fidler, R.~Urtasun, and A.~Yuille.
	\newblock Detect what you can: Detecting and representing objects using
	holistic models and body parts.
	\newblock In {\em Proc. {CVPR}}, 2014.
	
	\bibitem{msra}
	M.-M. Cheng, N.~J. Mitra, X.~Huang, P.~H.~S. Torr, and S.-M. Hu.
	\newblock Global contrast-based salient region detection.
	\newblock {\em {PAMI}}, 2015.
	
	\bibitem{CimpoiMKV16}
	M.~Cimpoi, S.~Maji, I.~Kokkinos, and A.~Vedaldi.
	\newblock Deep filter banks for texture recognition, description, and
	segmentation.
	\newblock {\em {IJCV}}, 2016.
	
	\bibitem{dai2015boxsup}
	J.~Dai, K.~He, and J.~Sun.
	\newblock Boxsup: Exploiting bounding boxes to supervise convolutional networks
	for semantic segmentation.
	\newblock In {\em Proc. {ICCV}}, 2015.
	
	\bibitem{DaiHS15}
	J.~Dai, K.~He, and J.~Sun.
	\newblock Instance-aware semantic segmentation via multi-task network cascades.
	\newblock In {\em Proc. {CVPR}}, 2016.
	
	\bibitem{DaiLHS16}
	J.~Dai, Y.~Li, K.~He, and J.~Sun.
	\newblock {R-FCN:} object detection via region-based fully convolutional
	networks.
	\newblock In {\em Proc. {NIPS}}, 2016.
	
	\bibitem{sfs}
	P.~Doll{\'{a}}r and C.~L. Zitnick.
	\newblock Fast edge detection using structured forests.
	\newblock {\em {PAMI}}, 2015.
	
	\bibitem{DongLHT16}
	C.~Dong, C.~C. Loy, K.~He, and X.~Tang.
	\newblock Image super-resolution using deep convolutional networks.
	\newblock {\em {PAMI}}, 2016.
	
	\bibitem{Eigen15}
	D.~Eigen and R.~Fergus.
	\newblock Predicting depth, surface normals and semantic labels with a common
	multi-scale convolutional architecture.
	\newblock In {\em Proc. {ICCV}}, 2015.
	
	\bibitem{pascal}
	M.~Everingham, S.~M.~A. Eslami, L.~J.~V. Gool, C.~K.~I. Williams, J.~M. Winn,
	and A.~Zisserman.
	\newblock The pascal visual object classes challenge: {A} retrospective.
	\newblock {\em {IJCV}}, 2015.
	
	\bibitem{farabet2013learning}
	C.~Farabet, C.~Couprie, L.~Najman, and Y.~LeCun.
	\newblock Learning hierarchical features for scene labeling.
	\newblock {\em {PAMI}}, 2013.
	
	\bibitem{ganin2014n}
	Y.~Ganin and V.~Lempitsky.
	\newblock N\^{} 4-fields: Neural network nearest neighbor fields for image
	transforms.
	\newblock In {\em Proc. {ACCV}}, 2014.
	
	\bibitem{GhiasiF16}
	G.~Ghiasi and C.~C. Fowlkes.
	\newblock Laplacian reconstruction and refinement for semantic segmentation.
	\newblock In {\em Proc. {ECCV}}, 2016.
	
	\bibitem{GidarisK16}
	S.~Gidaris and N.~Komodakis.
	\newblock Attend refine repeat: Active box proposal generation via in-out
	localization.
	\newblock {\em CoRR}, abs/1606.04446, 2016.
	
	\bibitem{girshick2014rcnn}
	R.~Girshick, J.~Donahue, T.~Darrell, and J.~Malik.
	\newblock Rich feature hierarchies for accurate object detection and semantic
	segmentation.
	\newblock In {\em CVPR}, 2014.
	
	\bibitem{Girshick15}
	R.~B. Girshick.
	\newblock Fast {R-CNN}.
	\newblock In {\em Proc. {ICCV}}, 2015.
	
	\bibitem{GkioxariGM15}
	G.~Gkioxari, R.~B. Girshick, and J.~Malik.
	\newblock Contextual action recognition with r*cnn.
	\newblock In {\em Proc. {ICCV}}, 2015.
	
	\bibitem{GruslysMDLG16}
	A.~Gruslys, R.~Munos, I.~Danihelka, M.~Lanctot, and A.~Graves.
	\newblock Memory-efficient backpropagation through time.
	\newblock {\em CoRR}, abs/1606.03401, 2016.
	
	\bibitem{HanLJSB15}
	X.~Han, T.~Leung, Y.~Jia, R.~Sukthankar, and A.~C. Berg.
	\newblock Matchnet: Unifying feature and metric learning for patch-based
	matching.
	\newblock In {\em Proc. {CVPR}}, 2015.
	
	\bibitem{hariharan2011semantic}
	B.~Hariharan, P.~Arbel{\'a}ez, L.~Bourdev, S.~Maji, and J.~Malik.
	\newblock Semantic contours from inverse detectors.
	\newblock In {\em Proc. {ICCV}}, 2011.
	
	\bibitem{hariharan2014simultaneous}
	B.~Hariharan, P.~Arbel{\'a}ez, R.~Girshick, and J.~Malik.
	\newblock Simultaneous detection and segmentation.
	\newblock In {\em Proc. {ECCV}}, 2014.
	
	\bibitem{hariharan2014hypercolumns}
	B.~Hariharan, P.~Arbel{\'a}ez, R.~Girshick, and J.~Malik.
	\newblock Hypercolumns for object segmentation and fine-grained localization.
	\newblock In {\em Proc. {CVPR}}, 2015.
	
	\bibitem{HarleyDK15}
	A.~W. Harley, K.~G. Derpanis, and I.~Kokkinos.
	\newblock Learning dense convolutional embeddings for semantic segmentation.
	\newblock {\em CoRR}, abs/1511.04377, 2015.
	
	\bibitem{HeZRS15}
	K.~He, X.~Zhang, S.~Ren, and J.~Sun.
	\newblock Deep residual learning for image recognition.
	\newblock In {\em Proc. {CVPR}}, 2016.
	
	\bibitem{iclrbnd}
	J.-J. Hwang and T.-L. Liu.
	\newblock Pixel-wise deep learning for contour detection.
	\newblock In {\em Proc. {ICLR}}, 2015.
	
	\bibitem{InsafutdinovPAA16}
	E.~Insafutdinov, L.~Pishchulin, B.~Andres, M.~Andriluka, and B.~Schiele.
	\newblock Deepercut: {A} deeper, stronger, and faster multi-person pose
	estimation model.
	\newblock {\em CoRR}, abs/1605.03170, 2016.
	
	\bibitem{batchnorm}
	S.~Ioffe and C.~Szegedy.
	\newblock Batch normalization: Accelerating deep network training by reducing
	internal covariate shift.
	\newblock In {\em Proc. {ICML}}, 2015.
	
	\bibitem{caffe}
	Y.~Jia, E.~Shelhamer, J.~Donahue, S.~Karayev, J.~Long, R.~B. Girshick,
	S.~Guadarrama, and T.~Darrell.
	\newblock Caffe: Convolutional architecture for fast feature embedding.
	\newblock In {\em Proceedings of the {ACM}}, 2014.
	
	\bibitem{JiangLYP13}
	P.~Jiang, H.~Ling, J.~Yu, and J.~Peng.
	\newblock Salient region detection by {UFO:} uniqueness, focusness and
	objectness.
	\newblock In {\em Proc. {ICCV}}, 2013.
	
	\bibitem{jacobs14}
	A.~Kanazawa, A.~Sharma, and D.~W. Jacobs.
	\newblock Locally scale-invariant convolutional neural networks.
	\newblock {\em CoRR}, abs/1412.5104, 2014.
	
	\bibitem{Keeler90}
	J.~D. Keeler, D.~E. Rumelhart, and W.~K. Leow.
	\newblock Integrated segmentation and recognition of hand-printed numerals.
	\newblock In {\em Proc. {NIPS}}, 1990.
	
	\bibitem{aistat}
	J.~J. Kivinen, C.~K.~I. Williams, and N.~Heess.
	\newblock Visual boundary prediction: {A} deep neural prediction network and
	quality dissection.
	\newblock In {\em AISTATS}, 2014.
	
	\bibitem{iclr16}
	I.~Kokkinos.
	\newblock Pushing the boundaries of boundary detection using deep learning.
	\newblock {\em ICLR}, 2016.
	
	\bibitem{KoMa05}
	I.~Kokkinos and P.~Maragos.
	\newblock An expectation maximization approach to the synergy between image
	segmentation and object categorization.
	\newblock In {\em Proc. {ICCV}}, volume~I, pages 617--624, 2005.
	
	\bibitem{krahenbuhl2011efficient}
	P.~Kr{\"a}henb{\"u}hl and V.~Koltun.
	\newblock Efficient inference in fully connected crfs with gaussian edge
	potentials.
	\newblock In {\em NIPS}, 2011.
	
	\bibitem{KrizhevskyNIPS2013}
	A.~Krizhevsky, I.~Sutskever, and G.~E. Hinton.
	\newblock Imagenet classification with deep convolutional neural networks.
	\newblock In {\em NIPS}, 2013.
	
	\bibitem{KTZ05}
	M.~P. Kumar, P.~Torr, and A.~Zisserman.
	\newblock {Obj-cut}.
	\newblock {\em Proc. {CVPR}}, 2005.
	
	\bibitem{LadickyZP14}
	L.~Ladicky, B.~Zeisl, and M.~Pollefeys.
	\newblock Discriminatively trained dense surface normal estimation.
	\newblock In {\em Proc. {ECCV}}, 2014.
	
	\bibitem{LarssonMS16}
	G.~Larsson, M.~Maire, and G.~Shakhnarovich.
	\newblock Learning representations for automatic colorization.
	\newblock {\em CoRR}, abs/1603.06668, 2016.
	
	\bibitem{LeCun1998}
	Y.~LeCun, L.~Bottou, Y.~Bengio, and P.~Haffner.
	\newblock {Gradient-based learning applied to document recognition}.
	\newblock In {\em Proc. IEEE}, 1998.
	
	\bibitem{LiY15}
	G.~Li and Y.~Yu.
	\newblock Visual saliency based on multiscale deep features.
	\newblock In {\em Proc. {CVPR}}, 2015.
	
	\bibitem{saliencycvpr16}
	G.~Li and Y.~Yu.
	\newblock Deep contrast learning for salient object detection.
	\newblock In {\em Proc. {CVPR}}, 2016.
	
	\bibitem{LiHKRY14}
	Y.~Li, X.~Hou, C.~Koch, J.~M. Rehg, and A.~L. Yuille.
	\newblock The secrets of salient object segmentation.
	\newblock In {\em Proc. {CVPR}}, 2014.
	
	\bibitem{LiangSFLY16}
	X.~Liang, X.~Shen, J.~Feng, L.~Lin, and S.~Yan.
	\newblock Semantic object parsing with graph {LSTM}.
	\newblock In {\em Proc. {CVPR}}, 2016.
	
	\bibitem{Adelaide}
	G.~Lin, C.~Shen, I.~D. Reid, and A.~van~den Hengel.
	\newblock Efficient piecewise training of deep structured models for semantic
	segmentation.
	\newblock {\em CVPR}, 2016.
	
	\bibitem{mscoco}
	T.~Lin, M.~Maire, S.~J. Belongie, J.~Hays, P.~Perona, D.~Ramanan,
	P.~Doll{\'{a}}r, and C.~L. Zitnick.
	\newblock Microsoft {COCO:} common objects in context.
	\newblock In {\em Proc. {ECCV}}, 2014.
	
	\bibitem{LiuSL14}
	F.~Liu, C.~Shen, and G.~Lin.
	\newblock Deep convolutional neural fields for depth estimation from a single
	image.
	\newblock In {\em Proc. {CVPR}}, 2015.
	
	\bibitem{LiuSLR15}
	F.~Liu, C.~Shen, G.~Lin, and I.~D. Reid.
	\newblock Learning depth from single monocular images using deep convolutional
	neural fields.
	\newblock {\em CoRR}, abs/1502.07411, 2015.
	
	\bibitem{ssd}
	W.~Liu, D.~Anguelov, D.~Erhan, C.~Szegedy, and S.~E. Reed.
	\newblock {SSD:} single shot multibox detector.
	\newblock {\em CoRR}, abs/1512.02325, 2015.
	
	\bibitem{parsenet}
	W.~Liu, A.~Rabinovich, and A.~C. Berg.
	\newblock Parsenet: Looking wider to see better.
	\newblock {\em CoRR}, abs/1506.04579, 2015.
	
	\bibitem{LongSD15}
	J.~Long, E.~Shelhamer, and T.~Darrell.
	\newblock Fully convolutional networks for semantic segmentation.
	\newblock In {\em Proc. {CVPR}}, 2015.
	
	\bibitem{MaireYP11}
	M.~Maire, S.~X. Yu, and P.~Perona.
	\newblock Object detection and segmentation from joint embedding of parts and
	pixels.
	\newblock In {\em ICCV}, 2011.
	
	\bibitem{MFTM01}
	D.~Martin, C.~Fowlkes, D.~Tal, and J.~Malik.
	\newblock A database of human segmented natural images and its application to
	evaluating segmentation algorithms and measuring ecological statistics.
	\newblock In {\em Proc. {ICCV}}, 2001.
	
	\bibitem{ishan}
	I.~Misra, A.~Shrivastava, A.~Gupta, and M.~Hebert.
	\newblock Cross-stitch networks for multi-task learning.
	\newblock In {\em Proc. {CVPR}}, 2016.
	
	\bibitem{context}
	R.~Mottaghi, X.~Chen, X.~Liu, N.-G. Cho, S.-W. Lee, S.~Fidler, R.~Urtasun, and
	A.~Yuille.
	\newblock The role of context for object detection and semantic segmentation in
	the wild.
	\newblock In {\em Proc. {CVPR}}, 2014.
	
	\bibitem{mumf94b}
	D.~Mumford.
	\newblock {Neuronal Architectures for Pattern Theoretic Problems}.
	\newblock In {\em Large Scale Theories of the Cortex}. MIT Press, 1994.
	
	\bibitem{NMY:ICCV:2015}
	T.~Narihira, M.~Maire, and S.~X. Yu.
	\newblock Direct intrinsics: Learning albedo-shading decomposition by
	convolutional regression.
	\newblock In {\em Proc. {ICCV}}, 2015.
	
	\bibitem{nyu}
	P.~K. Nathan~Silberman, Derek~Hoiem and R.~Fergus.
	\newblock Indoor segmentation and support inference from rgbd images.
	\newblock In {\em ECCV}, 2012.
	
	\bibitem{NewellYD16}
	A.~Newell, K.~Yang, and J.~Deng.
	\newblock Stacked hourglass networks for human pose estimation.
	\newblock {\em CoRR}, abs/1603.06937, 2016.
	
	\bibitem{NohHH15}
	H.~Noh, S.~Hong, and B.~Han.
	\newblock Learning deconvolution network for semantic segmentation.
	\newblock In {\em Proc. {ICCV}}, 2015.
	
	\bibitem{OquabBLS15}
	M.~Oquab, L.~Bottou, I.~Laptev, and J.~Sivic.
	\newblock Is object localization for free? - weakly-supervised learning with
	convolutional neural networks.
	\newblock In {\em Proc. {CVPR}}, 2015.
	
	\bibitem{papa15}
	G.~Papandreou, L.~Chen, K.~Murphy, and A.~L. Yuille.
	\newblock Weakly- and semi-supervised learning of a {DCNN} for semantic image
	segmentation.
	\newblock In {\em Proc. {ICCV}}, 2015.
	
	\bibitem{PapandreouKS15}
	G.~Papandreou, I.~Kokkinos, and P.~Savalle.
	\newblock Modeling local and global deformations in deep learning: Epitomic
	convolution, multiple instance learning, and sliding window detection.
	\newblock In {\em Proc. {CVPR}}, 2015.
	
	\bibitem{PerazziKPH12}
	F.~Perazzi, P.~Kr{\"{a}}henb{\"{u}}hl, Y.~Pritch, and A.~Hornung.
	\newblock Saliency filters: Contrast based filtering for salient region
	detection.
	\newblock In {\em Proc. {CVPR}}, 2012.
	
	\bibitem{PfisterCZ15}
	T.~Pfister, J.~Charles, and A.~Zisserman.
	\newblock Flowing convnets for human pose estimation in videos.
	\newblock In {\em Proc. {ICCV}}, 2015.
	
	\bibitem{PinheiroCD15}
	P.~H.~O. Pinheiro, R.~Collobert, and P.~Doll{\'{a}}r.
	\newblock Learning to segment object candidates.
	\newblock In {\em Proc. {NIPS}}, 2015.
	
	\bibitem{QinLXW15}
	Y.~Qin, H.~Lu, Y.~Xu, and H.~Wang.
	\newblock Saliency detection via cellular automata.
	\newblock In {\em Proc. {CVPR}}, 2015.
	
	\bibitem{RanjanPC16}
	R.~Ranjan, V.~M. Patel, and R.~Chellappa.
	\newblock Hyperface: {A} deep multi-task learning framework for face detection,
	landmark localization, pose estimation, and gender recognition.
	\newblock {\em CoRR}, abs/1603.01249, 2016.
	
	\bibitem{RenHGS15}
	S.~Ren, K.~He, R.~B. Girshick, and J.~Sun.
	\newblock Faster {R-CNN:} towards real-time object detection with region
	proposal networks.
	\newblock In {\em Proc. {NIPS}}, 2015.
	
	\bibitem{Ren12}
	X.~Ren and L.~Bo.
	\newblock Discriminatively trained sparse code gradients for contour detection.
	\newblock In {\em Proc. {NIPS}}, 2012.
	
	\bibitem{RonnebergerFB15}
	O.~Ronneberger, P.~Fischer, and T.~Brox.
	\newblock U-net: Convolutional networks for biomedical image segmentation.
	\newblock In {\em Proc. MICCAI}, 2015.
	
	\bibitem{ILSVRC15}
	O.~Russakovsky, J.~Deng, H.~Su, J.~Krause, S.~Satheesh, S.~Ma, Z.~Huang,
	A.~Karpathy, A.~Khosla, M.~Bernstein, A.~C. Berg, and L.~Fei-Fei.
	\newblock {ImageNet Large Scale Visual Recognition Challenge}.
	\newblock {\em IJCV}, 2015.
	
	\bibitem{SEZM+14}
	P.~Sermanet, D.~Eigen, X.~Zhang, M.~Mathieu, R.~Fergus, and Y.~LeCun.
	\newblock Overfeat: Integrated recognition, localization and detection using
	convolutional networks.
	\newblock In {\em Proc. {ICLR}}, 2014.
	
	\bibitem{shen2015deepcontour}
	W.~Shen, X.~Wang, Y.~Wang, X.~Bai, and Z.~Zhang.
	\newblock Deepcontour: A deep convolutional feature learned by positive-sharing
	loss for contour detection.
	\newblock In {\em Proc. {CVPR}}, 2015.
	
	\bibitem{ShenZJWZB16}
	W.~Shen, K.~Zhao, Y.~Jiang, Y.~Wang, Z.~Zhang, and X.~Bai.
	\newblock Object skeleton extraction in natural images by fusing
	scale-associated deep side outputs.
	\newblock In {\em Proc. {CVPR}}, 2016.
	
	\bibitem{simo2015deepdesc}
	E.~Simo-Serra, E.~Trulls, L.~Ferraz, I.~Kokkinos, P.~Fua, and F.~Moreno-Noguer.
	\newblock Discriminative learning of deep convolutional feature point
	descriptors.
	\newblock {\em Proc. {ICCV}}, 2015.
	
	\bibitem{simonyan2014very}
	K.~Simonyan and A.~Zisserman.
	\newblock Very deep convolutional networks for large-scale image recognition.
	\newblock In {\em Proc. {ICLR}}, 2015.
	
	\bibitem{SzegedyLJSRAEVR14}
	C.~Szegedy, W.~Liu, Y.~Jia, P.~Sermanet, S.~Reed, D.~Anguelov, D.~Erhan,
	V.~Vanhoucke, and A.~Rabinovich.
	\newblock Going deeper with convolutions.
	\newblock In {\em Proc. {CVPR}}, 2015.
	
	\bibitem{blindspots}
	C.~Szegedy, W.~Zaremba, I.~Sutskever, J.~Bruna, D.~Erhan, I.~J. Goodfellow, and
	R.~Fergus.
	\newblock Intriguing properties of neural networks.
	\newblock {\em CoRR}, abs/1312.6199, 2013.
	
	\bibitem{ToshevS14}
	A.~Toshev and C.~Szegedy.
	\newblock Deeppose: Human pose estimation via deep neural networks.
	\newblock In {\em Proc. {CVPR}}, 2015.
	
	\bibitem{TsogkasK12}
	S.~Tsogkas and I.~Kokkinos.
	\newblock Learning-based symmetry detection in natural images.
	\newblock In {\em Proc. {ECCV}}, 2012.
	
	\bibitem{TsogkasSemanticPart15}
	S.~Tsogkas, I.~Kokkinos, G.~Papandreou, and A.~Vedaldi.
	\newblock Deep learning for semantic part segmentation with high-level
	guidance.
	\newblock {\em CoRR}, abs/1505.02438, 2015.
	
	\bibitem{TCYZ03}
	Z.~W. Tu, X.~Chen, A.~Yuille, and S.~C. Zhu.
	\newblock {Image Parsing: Unifying Segmentation, Detection, and Recognition}.
	\newblock In {\em Proc. {ICCV}}, 2003.
	
	\bibitem{UijlingsF15}
	J.~R.~R. Uijlings and V.~Ferrari.
	\newblock Situational object boundary detection.
	\newblock In {\em Proc. {CVPR}}, 2015.
	
	\bibitem{WangLLLS15}
	K.~Wang, L.~Lin, J.~Lu, C.~Li, and K.~Shi.
	\newblock {PISA:} pixelwise image saliency by aggregating complementary
	appearance contrast measures with edge-preserving coherence.
	\newblock {\em IEEE Trans. Im. Proc.}, 2015.
	
	\bibitem{WangLRY15}
	L.~Wang, H.~Lu, X.~Ruan, and M.~Yang.
	\newblock Deep networks for saliency detection via local estimation and global
	search.
	\newblock In {\em Proc. {CVPR}}, 2015.
	
	\bibitem{WangFG15}
	X.~Wang, D.~F. Fouhey, and A.~Gupta.
	\newblock Designing deep networks for surface normal estimation.
	\newblock In {\em Proc. {CVPR}}, 2015.
	
	\bibitem{WeiRKS16}
	S.~Wei, V.~Ramakrishna, T.~Kanade, and Y.~Sheikh.
	\newblock Convolutional pose machines.
	\newblock In {\em Proc. {CVPR}}, 2016.
	
	\bibitem{witkin}
	A.~P. Witkin.
	\newblock Scale-space filtering.
	\newblock In {\em IJCAI}, 1983.
	
	\bibitem{HAZN}
	F.~Xia, P.~Wang, L.~Chen, and A.~L. Yuille.
	\newblock Zoom better to see clearer: Human part segmentation with auto zoom
	net.
	\newblock In {\em Proc. {ECCV}}, 2016.
	
	\bibitem{hed}
	S.~Xie and Z.~Tu.
	\newblock Holistically-nested edge detection.
	\newblock In {\em Proc. {ICCV}}, 2015.
	
	\bibitem{yago16}
	T.~F. Yago~Vicente, M.~Hoai, and D.~Samaras.
	\newblock Noisy label recovery for shadow detection in unfamiliar domains.
	\newblock In {\em Proc. {CVPR}}, 2016.
	
	\bibitem{eccv16lift}
	K.~M. Yi, E.~Trulls, V.~Lepetit, and P.~Fua.
	\newblock Lift: Learned invariant feature transform.
	\newblock In {\em Proc. {ECCV}}, 2016.
	
	\bibitem{ZagoruykoK15}
	S.~Zagoruyko and N.~Komodakis.
	\newblock Learning to compare image patches via convolutional neural networks.
	\newblock In {\em Proc. {CVPR}}, 2015.
	
	\bibitem{ZhangLLT14}
	Z.~Zhang, P.~Luo, C.~C. Loy, and X.~Tang.
	\newblock Facial landmark detection by deep multi-task learning.
	\newblock In {\em Proc. {ECCV}}, 2014.
	
	\bibitem{ZhaoOLW15}
	R.~Zhao, W.~Ouyang, H.~Li, and X.~Wang.
	\newblock Saliency detection by multi-context deep learning.
	\newblock In {\em Proc. {CVPR}}, 2015.
	
	\bibitem{crfrnn}
	S.~Zheng, S.~Jayasumana, B.~Romera-Paredes, V.~Vineet, Z.~Su, D.~Du, C.~Huang,
	and P.~Torr.
	\newblock Conditional random fields as recurrent neural networks.
	\newblock In {\em Proc. {ICCV}}, 2015.
	
\end{thebibliography}
\newcommand{\noopsort}[1]{} \newcommand{\printfirst}[2]{#1}
\newcommand{\singleletter}[1]{#1} \newcommand{\switchargs}[2]{#2#1}

\end{document}